%% file: acl2020.tex
\documentclass[11pt,a4paper]{article}
\usepackage[dvipsnames]{xcolor}
\usepackage[hyperref]{acl2020}
\usepackage{times}
\usepackage{latexsym}
\usepackage{booktabs}
\usepackage{graphicx}
\usepackage{subcaption}
\usepackage{amsmath}
\usepackage{amssymb}
\usepackage{cleveref}
\usepackage{linguex-mod}
\usepackage[hang,flushmargin]{footmisc} 
\usepackage{xspace}

\graphicspath{{./figs/}}

\newcommand{\key}[1]{\textsc{#1}}

\newcommand{\textcite}[1]{\citet{#1}}

\renewcommand{\firstrefdash}{}

\usepackage{microtype}

\aclfinalcopy % Uncomment this line for the final submission
\def\aclpaperid{3179} %  Enter the acl Paper ID here

\newcommand\blfootnote[1]{%
  \begingroup
  \renewcommand\thefootnote{}\footnote{#1}%
  \addtocounter{footnote}{-1}%
  \endgroup
}

\newcommand{\bxs}{\textsc{BLLIP-xs}\xspace}
\newcommand{\bsm}{\textsc{BLLIP-sm}\xspace}
\newcommand{\bmd}{\textsc{BLLIP-md}\xspace}
\newcommand{\blg}{\textsc{BLLIP-lg}\xspace}

\title{A Systematic Assessment of Syntactic Generalization\\in Neural Language Models
}

\author{Jennifer Hu$^1$, Jon Gauthier$^1$, Peng Qian$^1$, Ethan Wilcox$^2$, \and Roger P.\ Levy$^1$ \\
  $^1$Department of Brain and Cognitive Sciences, Massachusetts Institute of Technology \\
  $^2$Department of Linguistics, Harvard University\\
  \texttt{\{jennhu,pqian,rplevy\}@mit.edu}\\
  \texttt{jon@gauthiers.net}, \texttt{wilcoxeg@g.harvard.edu}} 

\date{}

\begin{document}
\setlength{\Exlabelwidth}{0.7em}
\setlength{\Exlabelsep}{0.7em}
\setlength{\SubExleftmargin}{1.3em}
\setlength{\Extopsep}{2pt}

\maketitle
\begin{abstract}

%As new state-of-the-art language models (LMs) continue to be released with impressive frequency, \todo{our methods of evaluating these models have largely remained the same}.\pengcomment{State-of-the-art neural models have achieved impressive perplexity results on major language modelling benchmarks, but it remains unclear whether optimizing for low perplexity leads to better syntactic generalization. Despite a growing body of works on targeted linguistic evaluation, we still lack systematic assessments due to  a lack of standardization across published test suites and controlled comparion across models.  Here we present a \textbf{scaled-up} and \textbf{controlled} ... } 

While state-of-the-art neural network models continue to achieve lower perplexity scores on language modeling benchmarks, it remains unknown whether optimizing for broad-coverage predictive performance leads to human-like syntactic knowledge. Furthermore, existing work has not provided a clear picture about the model properties required to produce proper syntactic generalizations. We present a systematic evaluation of the syntactic knowledge of neural language models, testing 20 combinations of model types and data sizes on a set of 34 English-language syntactic test suites.  We find substantial differences in syntactic generalization performance by model architecture, with sequential models underperforming other architectures. Factorially manipulating model architecture and training dataset size (1M--40M words), we find that variability in syntactic generalization performance is substantially greater by architecture than by dataset size for the corpora tested in our experiments. Our results also reveal a dissociation between perplexity and syntactic generalization performance.
%We find that model architecture clearly influences syntactic generalization performance: Transformer models and models with explicit hierarchical structure reliably outperform pure sequence models in their predictions. In contrast, we find no clear influence of the scale of training data on these syntactic generalization tests across tested models. We also find no clear relation between a model's perplexity and its syntactic generalization performance.
\blfootnote{Materials and code can be found at \url{https://github.com/cpllab/syntactic-generalization}.} %\todo{[cut last sentence?]} Our results demonstrate that sustained effort to scale up and carefully control targeted evaluation can lead to more robust and empirically-grounded understanding of recent gains made in natural language processing.

\end{abstract}

\section{Introduction} \label{sec:introduction}

A growing body of work advocates that assessment of neural language models should include both information-theoretic metrics, such as perplexity, as well as targeted linguistic evaluation. Benchmarks such as GLUE \citep{wang2019superglue,wang2018glue} have demonstrated that neural language models trained on naturalistic corpora for next-word prediction learn representations that can yield remarkable performance on many semantic tasks. Targeted syntactic evaluations have shown that these models also implicitly capture many \textbf{syntactic generalizations}, ranging from subject--verb agreement to  long-distance filler--gap dependencies \citep{Linzen:et-al:2016, Marvin:Linzen:2018,Futrell:et-al:2018,Wilcox:et-al:2019}.
% leave for related work section later
% In addition, probes into neural models' internal representations have demonstrated that they are capable of inducing a range of morphological and grammatical representations \cite{hewitt2019structural, belinkov2017neural}.
This paper aims to bring targeted evaluations of syntactic performance to scale, complementing similar developments in semantic evaluation \citep{mccoy2019right}.

Because the most widespread currency of evaluation for language models is perplexity---how well, on average, a model predicts a word in its context---a primary focus of this paper is the relationship between a model's perplexity and its performance on targeted syntactic evaluations. As 
%model 
perplexity improves, can we expect more human-like syntactic generalization?  How do training dataset size and model architecture jointly affect syntactic generalization?  And what picture of models' syntactic generalization emerges when evaluation is brought to scale, across dozens of controlled syntactic tests?

% ur understanding of what features of natural language syntax are implicitly captured by standard optimization of language models for perplexity.  Are lower-per

% First, these studies operate under the implicit assumption that syntactic generalization abilities are informative above and beyond perplexity. However, the empirical relationship between broad-coverage perplexity and SG performance has not been carefully tested. Second, because papers typically report performance on a handful of state-of-the-art models, there remains no controlled assessment of how different model design choices impact SG performance. Third, previous studies have focused on only a handful of syntactic evaluations at a time. 
% % have focused on one phenomenon of interest---or a cluster of inter-related phenomena as in \citet{marvin-linzen:2018-targeted}---there has been no assessment of SG variance across a broad range of syntactic patterns.

In this paper we offer initial answers to these questions, systematically assessing the syntactic generalization abilities of neural language models on 34 targeted test suites (33 adapted from previously published work, and 1 novel) covering a wide range of syntactic phenomena. Test suites are written using a standard format that allows for flexible predictions which more closely resemble those used in psycholinguistic studies, specifically allowing for predictions about interactions among multiple testing conditions. Performance on each test suite is reported as a Syntactic Generalization (SG) score. We group test suites into six syntactic circuits based on the linguistic representations needed to achieve high performance on each suite.

We train four classes of neural models and one baseline $n$-gram model on four datasets derived from a newswire corpus, consisting of 1, 5, 14, and 42 million tokens. While previous work has compared model architectures for a fixed dataset size \citep[e.g.][]{Wilcox:et-al:2019} and network sizes for a fixed architecture \citep[e.g.][]{van2019quantity}, our controlled regime allows us to make an apples-to-apples comparison across model architectures on a range of sizes. In addition, we evaluate several off-the-shelf models which were trained on datasets ranging up to 2 billion tokens.

Our results address the three questions posed above: First, for the range of model architectures and dataset sizes tested, we find a substantial dissociation between perplexity and SG score. %While off-the-shelf models do outperform medium data models on both perplexity and SG scores, we take the lack of relationship as evidence that, within similar training regimes, targeted evaluation of syntactic generalization offers complementary information about models' learning outcomes.
Second, we find a larger effect of model inductive bias than training data size on SG score, a result that accords with \citet{van2019quantity}. Models afforded explicit structural supervision during training outperform other models: One structurally supervised model is able to achieve the same SG scores as a purely sequence-based model trained on $\sim$100 times the number of tokens.
Furthermore, several Transformer models achieve the same SG score as a Transformer trained on $\sim$200 times the amount of data.
Third, we find that architectures have different relative advantages across types of syntactic tests, suggesting that the tested syntactic phenomena tap into different underlying processing capacities in the models. %that models' differing SG scores are due to variance on only a few key syntactic phenomena, specifically ones that deal with larger syntactic chunks.

\section{Background}

\subsection{Perplexity}

Standard language models are trained to predict the next token given a context of previous tokens.
%\footnote{More recent language models predict a token given both its left and right contexts, as in \citet{devlin2018bert,yang2019xlnet} 
%
Language models are typically assessed by their \emph{perplexity}, the inverse geometric mean of the joint probability of words $w_1,\dots,w_N$ in a held-out test corpus $C$:
\begin{align}
    \text{PPL}(C) &= p(w_1, w_2, \dots w_N)^{-\frac{1}{N}} %\nonumber \\
%    &= \left(\frac{1}{\prod_{i=1}^N p(w_i|w_1 \dots w_{i-1})}\right)^{\frac{1}{N}} \label{eq:ppl}
\end{align}

Models with improved perplexity have also been shown to better match various human behavioral measures, such as gaze duration during reading \citep{frank2011insensitivity,fossum2012sequential,goodkind2018predictive,Wilcox:et-al:2020:syntaxgym}. However, a broad-coverage metric such as perplexity may not be ideal for assessing human-like syntactic knowledge for a variety of reasons. %conflates the semantic and syntactic components of successful language modeling. 
In principle, a sentence can appear with vanishingly low probability but still be grammatically well-formed, such as \textit{Colorless green ideas sleep furiously} \citep{chomskysyntactic}.
%In practice, it has been shown that broad-coverage perplexity measures %do not reflect the inaccurate generalizations models may have made and
% may over- or underestimate the syntactic generalizations learned by a model. %\todo{For example, \citet{Tran:et-al:2018} find that Transformer models \citep{Vaswani:et-al:2017} may obtain lower perplexity on test sets than long short-term memory recurrent neural networks (LSTM RNNs) \citep{Hochreiter:Schmidhuber:1997}, but they make three times as many errors when tested on a simple agreement pattern between English subjects and verbs.} 
While perplexity remains an integral part of language model evaluation, fine-grained linguistic assessment can provide both more challenging and more interpretable tests to evaluate neural models. %language model performance.

\subsection{Targeted tests for syntactic generalization} \label{sec:targeted-eval}

Alternatively, a language model can be evaluated on its ability to make human-like generalizations for specific syntactic phenomena \citep{Linzen:et-al:2016,Lau:et-al:2017,Gulordava:et-al:2018}. The targeted syntactic evaluation paradigm \citep{Marvin:Linzen:2018,futrell-etal:2019-neural-language-models} incorporates methods from psycholinguistic experiments, designing sentences which hold most lexical and syntactic features of each sentence constant while minimally varying features that determine grammaticality or surprise characteristics of the sentence. For example, given the two strings \emph{The keys to the cabinet are on the table} and \emph{*The keys to the cabinet is on the table}, a model that has learned the proper subject--verb number agreement rules for English should assign a higher probability to the grammatical plural verb in the first sentence than to the ungrammatical singular verb in the second \citep{Linzen:et-al:2016}.

Although some targeted syntactic evaluations, such as the example discussed above, involve simple comparisons of conditional probabilities of a word in its context, other evaluations are more complex. We can demonstrate this with an evaluation of models' ``garden-pathing'' behavior \citep{futrell-etal:2019-neural-language-models}. For example, the sentence \emph{The child kicked in the chaos found her way back home} yields processing disruption for humans at the word \emph{found}.  This is because, up to right before that word, the part-of-speech ambiguous \emph{kicked} is preferentially interpreted as the main verb of the sentence, whereas it turns out to be a passive participle in a reduced relative clause modifying \emph{child}.  This garden-path disambiguation effect is ameliorated by replacing \emph{kicked} with \emph{forgotten}, which is not part-of-speech ambiguous (B below; \citealp{trueswell-etal:1994}) or by using an unreduced relative clause (C below; \citealp{ferreira-clifton:1986}).  In probabilistic language models, these garden-path disambiguation effects are well captured by word negative log probabilities, or \key{surprisals} \citep{hale2001probabilistic}: $S(w|C) = -\log_2 p(w|C)$, which are independently well-established to predict human incremental processing difficulty over several orders of magnitude in word probability \citep{Smith:Levy:2013}.  A targeted syntactic evaluation for garden-pathing is provided by comparing surprisals at the disambiguating word \emph{found} in the set of four examples below \citep{futrell-etal:2019-neural-language-models}:
\vspace{0.5em}

{\small

\renewcommand{\theExNo}{\Alph{ExNo}}%modified in version 4.0
\renewcommand{\Exarabic}{\Alph}%added in version 4.0

\ex. The child kicked in the chaos \textbf{found} \dots

\ex. The child forgotten in the chaos \textbf{found} \dots

\ex. The child who  was kicked in the chaos \textbf{found} \dots 

\ex. The child who was forgotten in the chaos \textbf{found} \dots

}
\vspace{0.5em}

\noindent
Successful human-like generalization involves three criteria: (i) \emph{found} should be less surprising (i.e., more probable) in B than A; (ii) \emph{found} should be more probable in C than A; (iii) the C--D surprisal difference should be smaller than the A--B surprisal difference---a $2\times 2$ \emph{interaction effect} on surprisal---because the syntactic disambiguation effect of not reducing the relative clause was achieved by using a part-of-speech unambiguous verb.

We will use these controlled tests to help us describe and test for human-like syntactic knowledge in language models.

\subsection{Related work}

The testing paradigm presented here differs in several crucial ways from recent, related syntactic assessments and provides complementary insights. %Unlike \citet{hale2019text}, which involves careful control of training data size, our approach tests models on minimally different experimental sentences, not fit to human EEG data. 
Unlike \citet{warstadt2019cola}, our approach does not involve fine-tuning, but rather assesses what syntactic knowledge is induced from the language modeling objective alone. The most closely related work is the Benchmark of Linguistic Minimal Pairs \citep{warstadt2019blimp}, which is a challenge set of automatically-generated sentence pairs also designed to test language models on a large set of syntactic phenomena. Our approach differs in important ways: we compare critical sentence regions instead of full-sentence probabilities, and employ a $2 \times 2$ paradigm with a strict, multi-fold success criterion inspired by psycholinguistics methodology. This allows us to factor out as many confounds as possible, such as the lexical frequency of individual tokens and low-level $n$-gram statistics. %By carefully controlling test items and separating them into test suites (see \Cref{sec:test-suites}), our objective is to provide an overall snapshot of syntactic performance as well as reveal structure-by-structure generalizations.

\section{Methods}

% The methods are really heavy, so we need some signposting methinks:
We designed a controlled paradigm for systematically testing the relationship between two design choices --- model class and dataset size --- and two performance metrics --- perplexity and syntactic generalization capacity. \Cref{sec:test-suites} describes the test suites collected for our evaluation, and \Cref{sec:data-preprocessing,sec:model-classes} describe the datasets and model classes investigated.

\subsection{Test suites} \label{sec:test-suites}

We assemble a large number of test suites inspired by the methodology of experimental sentence-processing and psycholinguistic research.  Each test suite contains a number of \key{items} (typically between 20 and 30), and each item appears in several \key{conditions}: across conditions, a given item will differ only according to a controlled manipulation designed to target a particular feature of grammatical knowledge. Each test suite contains at least one \key{prediction}, which specifies inequalities between surprisal values at pairs of regions/conditions that should hold if a model has learned the appropriate syntactic generalization. 

We expect language models which have learned the appropriate syntactic generalizations from their input to satisfy these inequalities without further fine-tuning. We compute accuracy on a test suite as the proportion of items for which the model's behavior conforms to the prediction. %
%
% In some previous studies, test suites consist of a single grammatical/ungrammatical contrast, such as the \textit{keys to the cabinet is/are} manipulation discussed in Section \ref{sec:targeted-eval}. One worry is that it might be possible for models to achieve very high accuracy for such manipulations using a na\"ive bag-of-words approach. To counter this, we formulate all of our test suites except for one as a 
%
Most of our test suites involve 2$\times$2 designs and a success criterion consisting of a conjunction of inequalities %among surprisals 
across conditions, as in the garden-pathing example described in Section~\ref{sec:targeted-eval}.\footnote{The exception is Center Embedding, which features a 2-condition design with a single-inequality criterion.} Random baseline accuracy varies by test suite and is $\sim$25\% overall. Most of these test suites and criteria are designed so that $n$-gram models cannot perform above chance for $n=5$  (sometimes greater).  

\paragraph{Syntactic coverage} In order to assess the coverage of our test suites, we manually inspected the phenomena covered in \citet{carnie2012syntax}, a standard introductory syntax textbook. Of the 47 empirical phenomena reviewed in the summary sections at the end of each chapter, our tests target 16 ($\sim$34\%). These are evenly distributed across the whole range of subject matter, with tests targeting phenomena in 11 of the 15 chapters ($\sim$73\%).\footnote{For more details on this analysis, see \Cref{sec:coverage}.}

\paragraph{Modifiers} Five test suites include paired modifier versions, where extra syntactically irrelevant (but semantically plausible) content, such as a prepositional phrase or relative clause, is inserted before the critical region being measured. We use these paired test suites to evaluate models' stability to intervening content within individual syntactic %generalization 
tests.

\paragraph{Circuits} The test suites are divided into 6 syntactic circuits, based on the type of algorithm required to successfully process each construction. We give a brief overview of each circuit below.\footnote{A full overview of our test suites is given in \Cref{sec:summary-test-suites}.}

\begin{itemize}
\item \textbf{Agreement} is a constraint on the feature values of two co-varying tokens. For example, the number feature of a verb must agree with the number feature of its upstream subject. We include 3 \textit{Subject-Verb Number Agreement} suites from \citet{Marvin:Linzen:2018}.
\item \textbf{Licensing} occurs when a particular token must exist within the scope of an upstream licensor token. Scope is determined by the tree-structural properties of the sentence. Test suites include  \textit{Negative Polarity Item Licensing (NPI)} (4 suites) and \textit{Reflexive Pronoun Licensing} (6 suites), both from \citet{Marvin:Linzen:2018}.
\item \textbf{Garden-Path Effects} are well-studied syntactic phenomena that result from tree-structural ambiguities that give rise to locally-coherent but globally implausible syntactic parses. Garden-path test suites include \textit{Main Verb / Reduced Relative Clause (MVRR)} (2 suites) and \textit{NP/Z Garden-paths (NPZ)} (4 suites), both from \citet{Futrell:et-al:2018}.
\item \textbf{Gross Syntactic Expectation} is a processor's expectation for large syntactic chunks such as verb phrases or sentences, and are often set up by subordinating conjunctions such as \textit{while}, \textit{although} and \textit{despite}. Our tests for gross syntactic expectation include \textit{Subordination} (4 suites) from \citet{Futrell:et-al:2018}.
\item \textbf{Center Embedding} sentences are sentences recursively nested within each other. Subject and verbs must match in a first-in-last-out order, meaning models must  approximate a stack-like data-structure in order to successfully process them. Our 2 suites of \textit{Center Embedding} sentences come from the items presented in \citet{wilcox-etal:2019-hierarchical-representation}.
\item \textbf{Long-Distance Dependencies} are co-variations between two tokens that span long distances in tree depth. Test suites include \textit{Filler-Gap Dependencies (FGD)} (6 suites) from \citet{wilcox-etal:2018-what-do-rnns} and \citet{Wilcox:et-al:2019}, and 2 novel \textit{Cleft} suites, described in detail below.
\end{itemize}

\paragraph{Novel test suite: Cleft} We introduce one novel test suite that assesses models' ability to process pseudo-cleft constructions, which are used to put a particular syntactic constituent into focus via passive transformation. Consider Example~\ref{ex:pseudo-cleft}:

\vspace{0.7em}
\setcounter{ExNo}{0}
\ex. \label{ex:pseudo-cleft}
\a. What he did after coming in from the rain was \textbf{eat a hot meal}. [DO/VP] \label{ex:VP-gram}
\b. *What he devoured after coming in from the rain was \textbf{eat a hot meal}. [LEX/VP] \label{ex:VP-ungram}
\c. *What he did after coming in from the rain was \textbf{a hot meal}. [DO/NP] \label{ex:NP-ungram}
\d. What he devoured after coming in from the rain was \textbf{a hot meal}. [LEX/NP] \label{ex:NP-gram}
\vspace{0.7em}

When this constituent is a verb, it must be replaced in the wh-clause that heads the sentence with the DO verb, as in \ref{ex:VP-gram}, below. However, when it is a noun, the lexical verb for which it serves as an object must be preserved, as in \ref{ex:NP-gram}. If models have properly learned the pseudo-cleft construction, then DO verbs should set up expectations for VPs (the region in bold should have a lower surprisal in \ref{ex:VP-gram} than in \ref{ex:VP-ungram}) and lexicalized verbs should set up expectations for NPs (the region in bold should have a lower surprisal in \ref{ex:NP-gram} than in \ref{ex:NP-ungram}). 

\subsection{Model training data}
\label{sec:data-preprocessing}

\begin{table}[t]
    \centering
    \resizebox{\linewidth}{!}{
    \begin{tabular}{lrrrr} \toprule
        BLLIP sizes: & \multicolumn{1}{c}{\sc xs} & \multicolumn{1}{c}{\sc sm} & \multicolumn{1}{c}{\sc md} & \multicolumn{1}{c}{\sc lg} \\ \midrule
        \# sentences & 40K & 200K & 600K & 1.8M \\
        \# tokens & 1M & 4.8M & 14M & 42M \\
        \# non-UNK types & 24K & 57K & 100K & 170K \\
        \# UNK types & 68 & 70 & 71 & 74 \\ \bottomrule
        % \# sentences & 39,908 & 199,364 & 598,480 & 1,755,715 \\
        % \# tokens & 958,678 & 4,771,737 & 14,352,978 & 42,415,926 \\
        % \# non-UNK types & 23,974 & 57,259 & 99,705 & 169,655 \\
        % \# UNK types & 68 & 70 & 71 & 74 \\ \bottomrule
    \end{tabular}
    }
    \caption{Statistics of training set for each corpus size.}
    \label{tab:corpora}
\end{table}

% \begin{table}[t]
% \footnotesize
%     \centering
%     \begin{tabular}{lr} \toprule
%         Model & \# training tokens \\ \midrule
%         GPT-2/GPT2-LG & 40GB \\
%         GRNN & 90 million\\ 
%         JRNN & 800 million\\
%         Transformer-XL & 103 million\\\bottomrule
%     \end{tabular}
%     \caption{Statistics of training data for off-the-shelf models.}
%     \label{tab:corpora}
% \end{table}

\paragraph{Corpora} %Perplexity scores depend not only on the quality of the model, but also on intrinsic properties of the corpus and data preprocessing. %Different corpora may have different upper/lower bounds on perplexity that are independent of the properties of the model. 
%For example, a corpus of fantasy fiction (where nonce or low-frequency words are common) may be harder to predict than a corpus of news articles about American politics (where a small set of entities are mentioned over and over again). 

We train and evaluate models on English newswire corpora of four different sizes, obtained by randomly sampling sections from the Brown Laboratory for Linguistic Information Processing 1987-89 Corpus Release 1 \citep[BLLIP;][]{BLLIP}. The corpora are sampled such that the training set of each corpus is a proper subset of each larger corpus. We call these four corpora \bxs (40K sentences, 1M tokens); \bsm (200K sentences, 5M tokens); \bmd (600K sentences, 14M tokens); and \blg (2M sentences, 42M tokens). Table~\ref{tab:corpora} summarizes statistics of the training set for each corpus.

To ensure consistency in perplexity evaluation across datasets, we report perplexity scores achieved by the models on a shared held-out test set. We additionally use a shared held-out validation for tuning and early stopping.

We use the NLTK implementation of the Penn Treebank tokenizer to process all datasets \citep{bird-loper-2004-nltk,Marcus:et-al:1993}.

\paragraph{Out-of-vocabulary tokens}
%Handling out-of-vocabulary tokens (OOVs) consistently across datasets is crucial for a meaningful perplexity comparison. 
For each corpus, we designate a token as OOV if the token appears fewer than two times in the training set. Our larger training datasets thus contain larger vocabularies than our smaller training datasets. This allows larger-training-set models to learn richer word-specific information, but may also harm perplexity evaluation because they have vocabulary items that are guaranteed to not appear in the \bxs test set. This means that perplexity scores across training dataset sizes will not be strictly comparable: if a larger-training-set model does better than a smaller-training-set model, we can be confident that it has meaningfully lower perplexity, but the reverse is not necessarily the case. The exception to the above is GPT-2, which uses sub-words from byte-pair encoding and has no OOVs (see also Footnote~\ref{fn:gpt2}).

\paragraph{Unkification}
%Like coarse tokenization, aggressive unkification gives models a perplexity advantage. In the extreme case, if all words are mapped to UNK, then any model will trivially achieve perplexity of 0. % Using a set of fine-grained UNK tokens (e.g.\ \texttt{UNK-ed}, \texttt{UNK-ly}, \dots) puts a model somewhere in between no \texttt{unk}ification and single-class unkification (i.e.\ mapping all OOV tokens to a single UNK token).

We follow the convention used by the Berkeley parser \citep{petrov-klein-2007-improved}, which maps OOVs to UNK classes which preserve fine-grained information such as orthographic case distinctions and morphological suffixes (e.g.\ \texttt{UNK-ed}, \texttt{UNK-ly}). Before training, we verified that the UNK classes in the test and validation sets were all present in the training set.

\subsection{Model classes}
\label{sec:model-classes}

\begin{table}[t]
    \centering
    \resizebox{\linewidth}{!}{
    \begin{tabular}{lcccc} \toprule
         & \# layers & \# hidden units & Embedding size \\ \midrule
        LSTM & 2 & 256 & 256  \\
        ON-LSTM & 3 & 1150 & 400 \\
        % ON-LSTM & 2 & 256 & 256 & 500 \\
        RNNG & 2 & 256 & 256 \\
        GPT-2 & 12 & 768 & 768 \\ \bottomrule
        % Transfo. & 2 & 256 & 256 & 500 \\ \bottomrule
        % \midrule
        % \textsc{Off-the-shelf models} & & & \\
        % GRNN & & 650 & \\
        % JRNN & & & \\
        % Transformer-XL & & & \\
        % GPT-2 & 48 & & \\
        % GPT-2-XL & & & \\ \bottomrule
    \end{tabular}}
    \caption{Size of neural models in our controlled experiments.} %\todo{ON weird dropout, lr, Transformer 16 heads}}
    \label{tab:models}
\end{table}

\begin{table}[t]
    \centering
    \resizebox{\linewidth}{!}{
    \begin{tabular}{lrrrr} \toprule
       BLLIP sizes: & \multicolumn{1}{c}{\sc xs} & \multicolumn{1}{c}{\sc sm} & \multicolumn{1}{c}{\sc md} & \multicolumn{1}{c}{\sc lg} \\
        \midrule
        LSTM & 13.4M & 30.5M & 52.2M & 88.1M \\
        ON-LSTM & 30.8M & 44.2M & 61.2M & 89.2M \\
        RNNG & 22.8M & 48.4M & 81.1M & 134.9M \\
        GPT-2 & 124.4M & 124.4M & 124.4M & 124.4M \\ \bottomrule
%        LSTM & \bxs & 13,386,731 \\
%        LSTM & \bsm & 30,462,962 \\
%        LSTM & \bmd & 52,238,273 \\
%        LSTM & \blg & 88,124,162 \\ \midrule
%        ON-LSTM & \bmd & 61,232,757 \\
%        ON-LSTM & \blg & 89,283,910 \\ \midrule
%        RNNG & \bxs & 22,830,370 \\
%        RNNG & \bsm & 48,428,328 \\
%        RNNG & \bmd & 81,070,071 \\
%        RNNG & \blg & 134,863,928 \\ \midrule
%        GPT-2 & all & 124,439,808 \\ \bottomrule
    \end{tabular}
    }
    \caption{Parameter counts for neural models in our controlled experiments.
    \label{tab:param-counts}}
\end{table}

\begin{table}[t]
    \centering
    \resizebox{\linewidth}{!}{
    \begin{tabular}{lrrrr} \toprule
        BLLIP sizes: & \multicolumn{1}{c}{\sc xs} & \multicolumn{1}{c}{\sc sm} & \multicolumn{1}{c}{\sc md} & \multicolumn{1}{c}{\sc lg} \\ \midrule
        LSTM & 98.19 & 65.52 & 59.05 & 57.09 \\
        ON-LSTM & 71.76 & 54.00 & 56.37 & 56.38 \\
        RNNG & 122.46 & 86.72 & 71.12 & 69.57 \\ 
        GPT-2 & 529.90 & 183.10 & 37.04 & 32.14 \\ 
        $n$-gram & 240.21 & 158.60 & 125.58 & 106.09 \\ \bottomrule
    \end{tabular}
    }
    \caption{Perplexity averages achieved by each controlled model on each corpus. Perplexity scores across training dataset sizes are not always strictly comparable (see Section~\ref{sec:data-preprocessing}).}
    \label{tab:ppl}
\end{table}

In order to study the effects of model inductive bias and dataset size, we trained a fleet of models with varying inductive biases on each corpus. Because many of our test suites exploit ambiguities that arise from incremental processing, we restrict evaluation to left-to-right language models; future work could involve evaluation of bidirectional models \citep{Devlin:et-al:2018,yang2019xlnet} on an appropriate subset of our test suites, and/or adaptation of our suites for use with bidirectional models \citep{goldberg2019assessing}. Training ran until convergence of perplexity on a held-out validation set. Wherever possible, we trained multiple seeds of each model class and corpus size. We use the model sizes and training hyperparameters reported in the papers introducing each model (\Cref{tab:models}).\footnote{Due to computational constraints, we performed only minimal tuning past these recommended hyperparameters.} The full parameter counts and perplexity scores for each model $\times$ corpus combination are given in Tables~\ref{tab:param-counts} and~\ref{tab:ppl}, respectively.

\paragraph{LSTM} Our baseline neural model is a vanilla long short-term memory network \citep[LSTM;][]{Hochreiter:Schmidhuber:1997} based on the boilerplate PyTorch implementation \citep{paszke2017automatic}.

\paragraph{Ordered-Neurons} We consider the Ordered-Neurons LSTM architecture \citep[ON-LSTM;][]{Shen:et-al:2019}, which encodes an explicit bias towards modeling hierarchical structure.

\paragraph{RNNG} Recurrent neural network grammars \citep[RNNG;][]{Dyer:et-al:2016} model the joint probability of a sequence of words and its syntactic structure.
    % combining a symbolic approach with neural network language modeling. %\todo{hyperparameters: D0.3-2-256-256-16-256} 
RNNG requires labeled trees that contain complete constituency parses, which we produce for BLLIP sentences with an off-the-shelf constituency parser \citep{kitaev-klein-2018-constituency}.\footnote{While the BLLIP corpus already contains Treebank-style parses, we strip the terminals and re-parse in order to obtain more accurate, up-to-date syntactic parses.} %To our knowledge, ours is the largest RNNG model reported in the literature, measured in terms of training data size. 
To compute surprisals from RNNG, we use word-synchronous beam search \citep{stern2017effective} to approximate the conditional probability of the current word given the context.

\paragraph{Transformer} Transformer models \citep{vaswani2017attention} have recently gained popularity in language processing tasks. We use GPT-2 \citep{radford2019language} as a representative Transformer model and train it from scratch on our BLLIP corpora.\footnote{Our GPT-2 code is based on \href{https://github.com/nshepperd/gpt-2}{\texttt{nshepperd/gpt-2}}. The model vocabulary consists of byte-pair encoded sub-words extracted from the GPT-2 pre-trained model, not from the BLLIP training corpora.  To calculate GPT-2 perplexities, we divide the sum of all sub-word conditional log-probabilities by the total number of words in the corpus. \label{fn:gpt2}}

%We note that these GPT-2 models were trained using byte pair encoding (BPE), which may conflate model class with properties of the dataset. We will revisit this issue in \Cref{sec:bpe}.

\paragraph{$n$-gram} As a baseline, we consider a 5-gram model with modified Kneser-Ney smoothing.

\subsection{Off-the-shelf models} We also test five off-the-shelf models: GRNN, trained on 90M tokens from Wikipedia \citep{Gulordava:et-al:2018}; JRNN, trained on 800M tokens from the 1 Billion Word Benchmark \citep{Jozefowicz:et-al:2016}; Transformer-XL, trained on 103M tokens from WikiText-103 \citep{Dai:et-al:2019}; and the pre-trained GPT-2 and GPT-2-XL, trained on 40GB of web text \citep{radford2019language}. These models are orders of magnitude larger than our controlled ones in parameter count and/or training set size.

%  We do not report perplexity scores for the off-the-shelf models, as they have been trained with a variety of vocabularies and tokenization policies.

\begin{figure}[t]
    \centering
    \includegraphics[width=\linewidth]{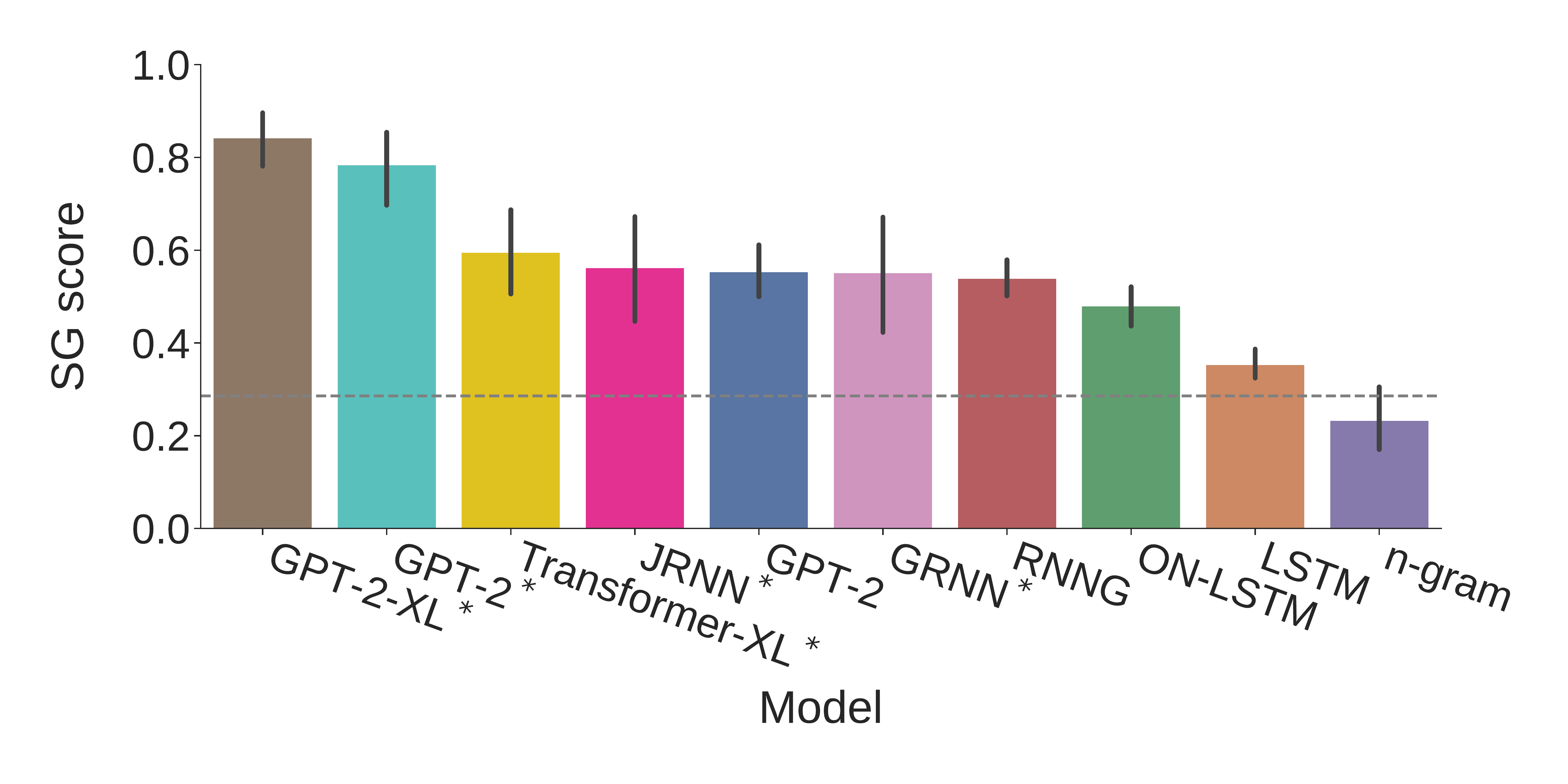}
    \caption{Average SG score by model class. Asterisks denote off-the-shelf models. Error bars denote bootstrapped 95\% confidence intervals of the mean.}
    \label{fig:sg-model}
\end{figure}

\section{Results}

\Cref{fig:sg-model} shows the average accuracy of all models on the complete set of SG test suites. Asterisks denote off-the-shelf models. All neural models achieve a SG score significantly greater than a random baseline (dashed line). However, the range within neural models is notable, with the best-performing model (GPT-2-XL) scoring over twice as high as the worst-performing model (LSTM). Also notable are the controlled GPT-2 and RNNG models, which achieve comparable performance to Transformer-XL and JRNN, despite being trained on significantly smaller data sizes.

We now return to the three major issues presented in Section \ref{sec:introduction}. In \ref{sec:results-SG-perplexity} we present evidence that SG score is dissociated from perplexity. In \ref{sec:results-inductive-bias} we argue that model architecture accounts for larger gains in SG score than amount of training data. And in \ref{sec:results-circuit} we show that this cross-architecture difference is due largely to variance on a handful of key test suites.

\subsection{Syntactic generalization and perplexity} \label{sec:results-SG-perplexity}

\begin{figure}
    \centering
    \includegraphics[width=\linewidth]{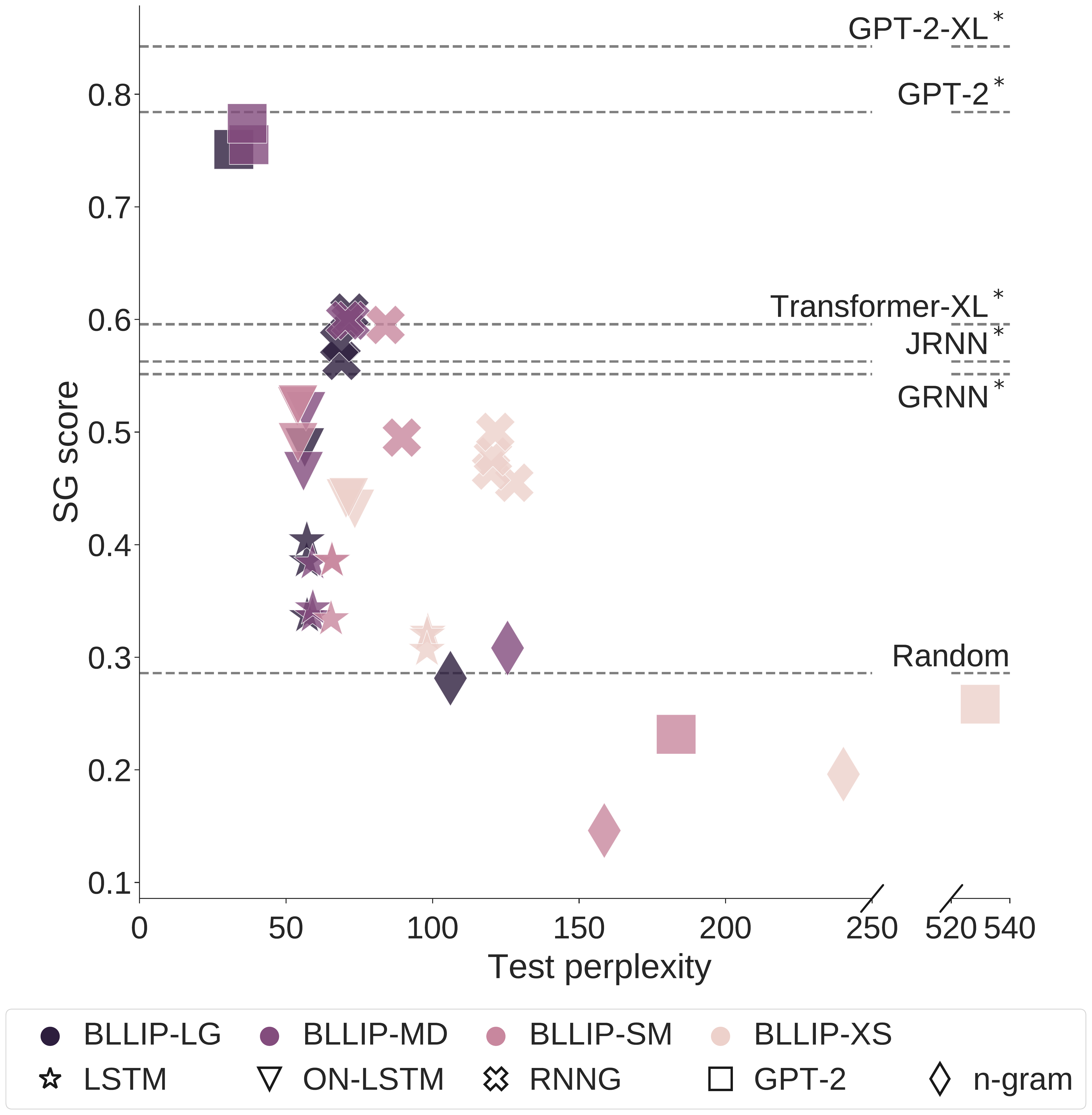}
    \caption{Relationship between SG score and perplexity on our held-out BLLIP test set for each model.}
    \label{fig:sg-perplexity}
\end{figure}

\begin{figure*}[t]
    \centering
    \includegraphics[width=\linewidth]{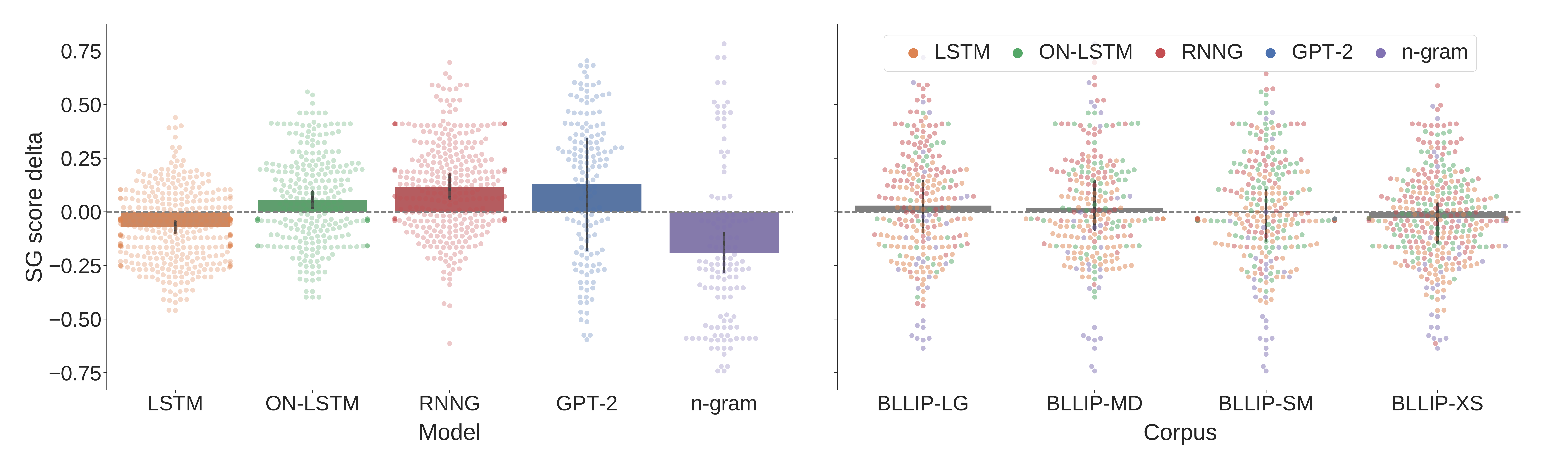}
    \caption{Main results of our controlled evaluation of model class and dataset size. SG score varies more by model class (left) than by training dataset size (right).%\todo{Left: Model class has a strong effect on SG score. Right: Data scale has little effect on SG score.}
    }
    \label{fig:sg-controlled}
\end{figure*}

% FIGURE 4 USED TO BE HERE

\Cref{fig:sg-perplexity} shows the relationship between SG score and perplexity on the BLLIP test set across models and training set sizes. %In addition to marking the performance of our controlled models, dashed lines to show the SG score of five off-the-shelf models.\footnote{We do not report their perplexity, as they have been trained with differently-sized vocabularies and tokenization policies.} 
As expected, $n$-gram models never rise appreciably above chance in SG score. Among neural models, GPT-2 achieves both the worst (\bxs and \bsm) and best (\bmd and \blg) performance; the impressive performance of these latter models %obtained in training with our larger BLLIP models 
comes with the caveat that the sub-words come from the pre-trained GPT-2 model, tacitly importing information from a larger training dataset (see further discussion in Section~\ref{sec:bpe}).
%models lies at the upper left and bottom right corners of the figure, outside of a large cluster of all other models. We see both of these models as exceptions to a broader trend. 
%\todo{By design, these n-gram models perform at or below chance on our test suites. The GPT-2 models trained on the smaller BLLIP corpora perform at chance, and those trained on \blg and \bmd achieve the best performance in both SG score and perplexity of all our models. However, this unusual behavior may be due to the pre-processing scheme used by GPT-2; we discuss this further in \Cref{sec:bpe}.}
%\footnote{We discuss this in greater detail in Section~\ref{sec:bpe}.}} 
%\todo{Say why GPT is exceptional!}
%Turning to the other models and comparing within corpus (same color in the figure),\footnote{As discussed earlier, models trained on differently sized corpora cannot be strictly compared in perplexity due to differing vocabulary sizes. Thus we cannot fairly evaluate the relationship between perplexity and SG score \emph{across} corpus sizes.} we see 
For the remaining neural models, there is no simple relationship between perplexity and SG score, especially once training dataset size is controlled for (comparing points in Figure~\ref{fig:sg-perplexity} of the same color). For example, there is a remarkable amount of variance in the SG score of models trained on \blg not explained by perplexity. This suggests that targeted syntactic evaluation can reveal information that may be orthogonal to perplexity. %\footnote{The apparent exception to this generalization may be the controlled GPT-2 models: for each BLLIP training corpus, the GPT-2 architecture is either best or worst among all neural models on \emph{both} SG score and perplexity. %However, this may be due to the pre-processing scheme used by GPT-2. %However, these models are trained on a representation of BLLIP using sub-words from the pre-trained GPT-2 model, tacitly importing information from a larger training dataset. 

\subsection{Inductive bias and data scale} \label{sec:results-inductive-bias}

% \begin{figure*}[t]
%     \centering
%     \includegraphics[width=\linewidth]{figs/controlled.pdf}
%     \caption{Main results of our controlled evaluation of model class and dataset size. Left: Model class has a strong effect on SG score. Right: Data scale has little effect on SG score.}
%     \label{fig:sg-controlled}
% \end{figure*}

% \begin{figure*}[t]
%     \centering
%     \includegraphics[width=\linewidth]{figs/controlled_circuit.pdf}
%     \caption{Controlled evaluation results, split across test suite circuits. Left: differences in model class induce significant differences in SG scores for several circuits. Right: differences in training data size do not reliably account for differences in SG score for any circuit.}
%     \label{fig:sg-controlled-circuit}
% \end{figure*}

In order to decouple the effects of model class and data scale from test suite difficulty, we represent a particular trained model's performance on each test suite as a delta relative to the average performance of all models on this test suite. Unless noted otherwise, the remainder of the figures in this section plot a score delta, aggregating these deltas within model classes or corpus types.

\Cref{fig:sg-controlled} tracks the influence of model class and data scale across the model types tested in our experiments, with SG score deltas on the y-axis. The left-hand panel shows the difference in SG score by model class. We find that model class clearly influences SG score: for example, the error bars (bootstrapped 95\% confidence intervals of the mean) for RNNG and LSTM do not overlap. 
%the error bars (bootstrapped 95\% confidence intervals of the mean) for the best model (RNNG) and the worst neural model (LSTM) do not overlap. 
The right-hand panel shows the difference in SG score delta by training dataset, and shows a much more minor increase in mean SG score as training data increases.

We tested the influence of these factors quantitatively using a linear mixed-effects regression model, predicting suite-level performance as a feature of model architecture and training dataset size (represented as log-number of words). Both features made statistically significant contributions to SG score (both $p < 0.001$).
%A linear mixed-effects regression model analysis of suite-level performance finds statistically significant effects of architecture and training dataset size (in log-number of words; both $p<0.001$).
However, predictor ablation indicates that architecture affects regression model fit more (AIC=--581 when dataset size is ablated; AIC=--574 when architecture is ablated).\footnote{$n$-grams and/or GPT-2 could arguably be expected to have qualitatively different sensitivity to training dataset size (the latter due to byte-pair encoding), so we repeated the analyses here and in Section~\ref{sec:results-circuit} excluding both architectures individually as well as simultaneously. In all cases the same qualitative patterns described in the main text hold.}

Beyond the above analysis, our GPT-2 results offer another striking example of the influence of model architecture relative to data scale. \Cref{fig:sg-perplexity} shows that our controlled \bmd and \blg GPT-2 models achieve roughly the same SG score as the pre-trained GPT-2 model, despite being trained on less than 1\% of the data used by the pre-trained model. %Furthermore, there is little difference between the SG scores achieved by the \bmd and \blg models, suggesting 
This suggests diminishing returns to training data scale for syntactic generalization performance.%\footnote{It is worth noting that the smaller GPT-2 models fare poorly in both perplexity and SG evaluations. This suggests an interaction between model class and data scale below some minimum data threshold for some models; GPT-2's sub-word vocabulary may be implicated in this case, as we discuss in Section~\ref{sec:data-preprocessing}.}

% \todo{The off-the-shelf models achieve quite high syntactic generalization accuracies---around 80\% on average, in the case of the pre-trained GPT-2 models. \Cref{fig:sg-perplexity} suggests that the strong performance of GPT-2 and related models is due more to architectural choices than to the scale of their training data: the GPT-2 models trained on \blg and \bmd achieve comparable performance to the pre-trained GPT-2 model. However, the GPT-2 models trained on \bsm and \bxs perform poorly in both perplexity and SG score, suggesting an interaction between dataset size and architecture that we leave to be explored in future work.}

% \subsection{Quantitative assessment} %of perplexity--syntactic performance relationship}
% \label{sec:quantitative-assessment}

% To assess the strength of evidence for 
% \footnote{\verb!accuracy ~ model/ppl + (ppl | model:suite) + (ppl | model:circuit)! and \verb!accuracy ~ training_set/ppl +  (ppl | training_set:suite) + (ppl || training_set:circuit)!}

% \begin{figure*}[t]
%     \centering
%     \includegraphics[width=\linewidth]{figs/allmodels_circuit.pdf}
%     \caption{Evaluation results on all models, split across test suite circuits.}
%     \label{fig:sg-allmodels-circuit}
% \end{figure*}

\begin{figure*}[t]
    \centering
    \includegraphics[width=\linewidth]{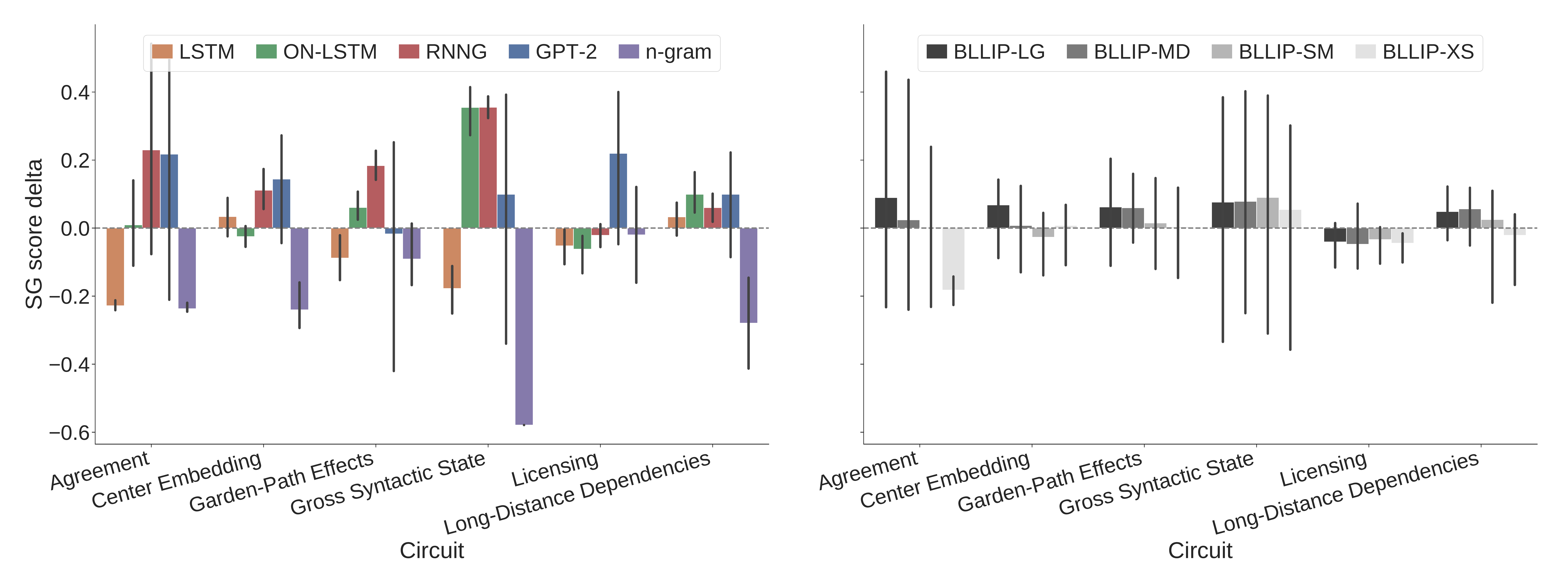}
    \caption{Controlled evaluation results, split across test suite circuits. Circuit-level differences in SG score vary more by model class (left) than by training dataset size (right).
    %\todo{Left: differences in model class induce significant differences in SG scores for several circuits. Right: differences in training data size do not reliably account for differences in SG score for any circuit.}
    }
    \label{fig:sg-controlled-circuit}
\end{figure*}

\begin{figure*}[t]
    \centering
    \includegraphics[width=\linewidth]{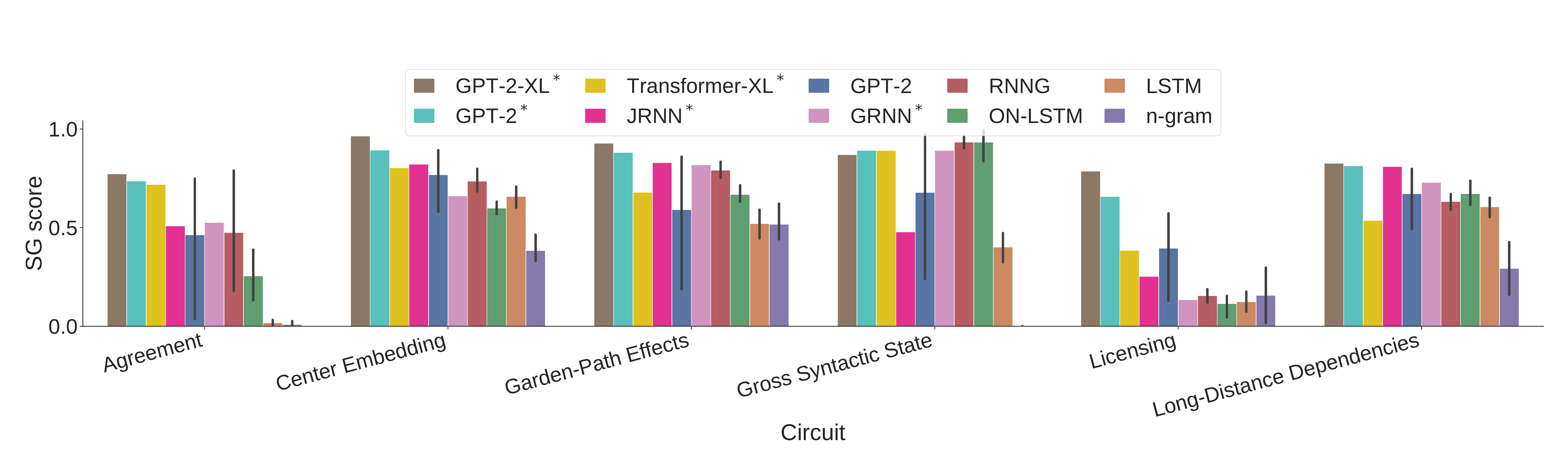}
    \caption{Evaluation results on all models, split across test suite circuits.}
    \label{fig:sg-allmodels-circuit}
\end{figure*}

\subsection{Circuit-level effects on SG score} \label{sec:results-circuit}

% \begin{figure*}[t]
%     \centering
%     \includegraphics[width=\linewidth]{figs/controlled_circuit.pdf}
%     \caption{Controlled evaluation results, split across test suite circuits. Left: differences in model class induce significant differences in SG scores for several circuits. Right: differences in training data size do not reliably account for differences in SG score for any circuit.}
%     \label{fig:sg-controlled-circuit}
% \end{figure*}

% \begin{figure*}[t]
%     \centering
%     \includegraphics[width=\linewidth]{figs/allmodels_circuit.pdf}
%     \caption{Evaluation results on all models, split across test suite circuits.}
%     \label{fig:sg-allmodels-circuit}
% \end{figure*}

% The left panel of \Cref{fig:sg-controlled-circuit} compares the delta scores of different models within test suite circuits. We see that inductive bias has a clear effect in some circuits --- Agreement and Gross Syntactic State, for example --- while other circuits are less affected by model choice. In contrast, the right panel of \Cref{fig:sg-controlled-circuit} shows that data scale has no clear effect on any within-suite delta score.

\Cref{fig:sg-controlled-circuit} shows the breakdown at the circuit level by model architecture (left) and training dataset size (right).
%with the same inductive bias trained on different datasets (right). 
The right panel demonstrates little effect of dataset size on SG score delta within most circuits, except for Agreement, on which the models trained on our smallest dataset fare poorly. In the left panel we find substantial between-circuit differences across architectures. Linear mixed-effects analyses support this finding: interactions with circuit are significant for both training dataset size and model architecture, but stronger for the latter (AIC=--654 and AIC=--623 when size and architecture are respectively ablated).    

While model inductive biases separate clearly in performance on some circuits, they have little effect on performance on Licensing. This minimally suggests that Licensing taps into a distinct syntactic process within language models. One potential explanation for this is that the interactions tested by Licensing involve tracking two co-varying tokens where the downstream token is optional \citep[see e.g.][]{hu-etal:2020-a-closer-look}. %We see the most substantial contrasts between models in the Garden-Path Effects and Gross Syntactic State circuits. These results suggest that much of the variance in SG score between models may rely on a model's ability to develop robust expectations about larger chunks of language---something which is enabled by the the explicit structural supervision of the RNNG and ON-LSTM.

We show the circuit-level breakdown of absolute SG scores for all models (including off-the-shelf) in \Cref{fig:sg-allmodels-circuit}. In general, the models that obtain high SG scores on average (as in \Cref{fig:sg-model}) also perform well across circuits: pre-trained GPT-2 and GPT-2-XL outperform all other models on each circuit, including Licensing, on which JRNN, GRNN, and most of our custom-trained models perform particularly poorly. Again, we highlight the impressive performance of RNNG: it achieves comparable average performance to GRNN on all circuits, despite being trained on a fraction of the data size.

%Of the nine significant differences between models of different architectures, six of them are found in the \textit{Gross Syntactic State} and \textit{Garden-Path Effects} circuits.

%\Cref{fig:sg-circuit-comparison} shows example correlations between Garden-Path Effects and three other circuits. \jenncomment{Ethan, can you explain the figure with some linguistic terminology?} Our scaled-up study lays the foundation for finding behavioral correlates of the localized internal representations used in syntactic tasks.

%\begin{figure*}[ht]
    %\centering
   % \includegraphics[width=\linewidth]{figs/correlations.pdf}
    %\caption{Example relationships between circuit-level performance. \textbf{Left:} poor relationship between Garden-Path Effects and Licensing. \textbf{Middle:} moderate relationship between Garden-Path Effects and Long-Distance Dependencies. \textbf{Right:} strong relationship between Garden-Path Effects and Gross Syntactic State.}
   % \label{fig:sg-circuit-comparison}
%\end{figure*}

% \jenncomment{Test suite stuff: (1) within-tag/circuit ppl-SG correlations, (2) circuit-circuit coordination heatmap, (3) robustness to stability modification}

% \todo{sensitivity to type vs token freq -- ppl vs SG}

\subsection{Stability to modifiers}

We separately investigate the degree to which models' syntactic generalizations are robustly stored in memory. For five test suites (Center Embedding, Cleft, MVRR, NPZ-Ambiguous, NPZ-Object), we designed minimally edited versions where syntactically irrelevant intervening content was inserted before the critical region. An ideal model should robustly represent syntactic features of its input 
across these modifier insertions.

\begin{figure}[ht]
    \centering
    \includegraphics[width=\linewidth]{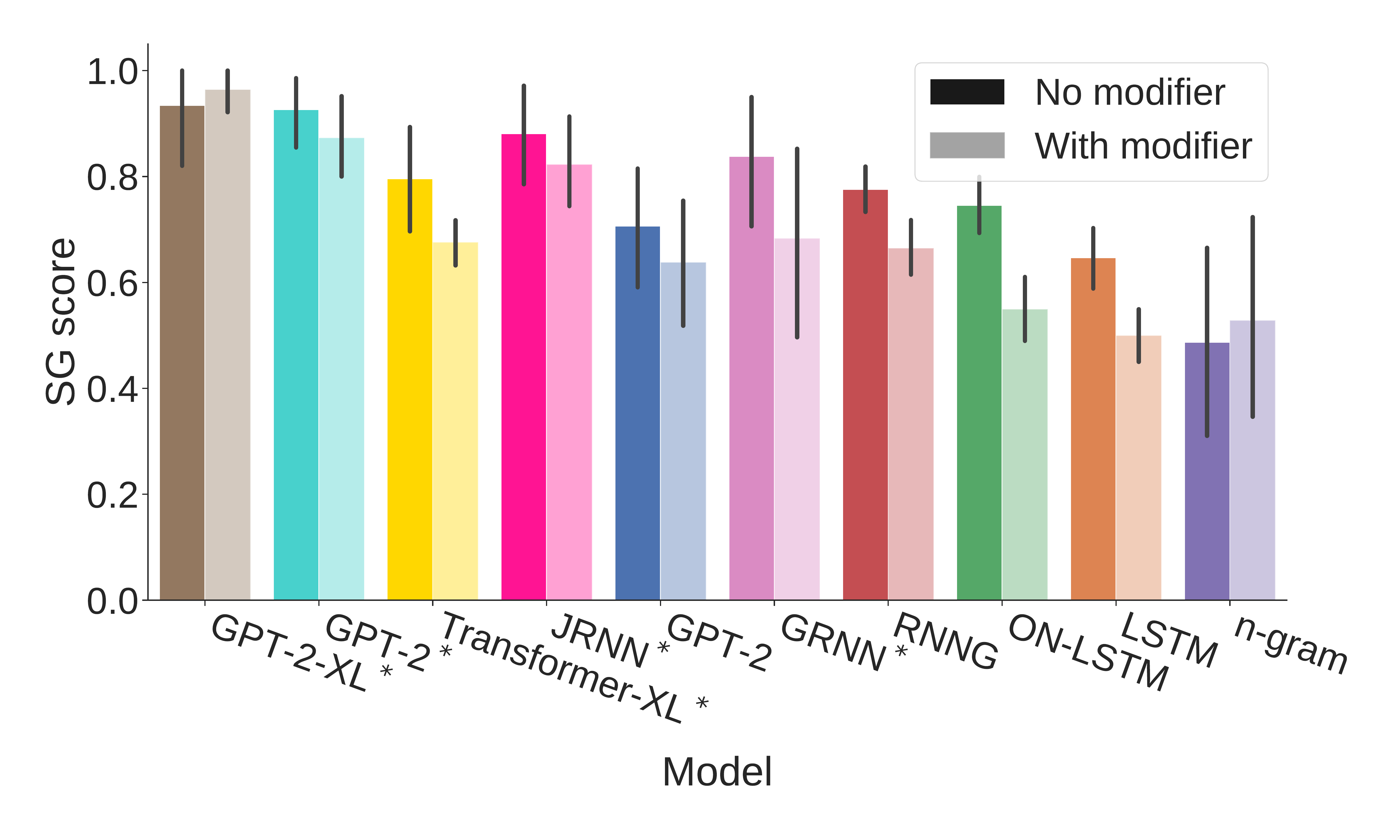}
    \caption{SG score on the pairs of test suites with and without intervening modifiers: Center Embedding, Cleft, MVRR, NPZ-Ambiguous, and NPZ-Object.}
    \label{fig:sg-stability}
\end{figure}

In \Cref{fig:sg-stability} we plot models' average scores on these five test suites (dark bars) and their minimally edited versions (light bars), evaluating how robust each model is to intervening %irrelevant 
content. Among models in our controlled experiments, we see that model class clearly influences the degree to which predictions are affected by intervening content (compare e.g.\ the stability of RNNG to that of ON-LSTM). Some off-the-shelf models, such as GPT-2-XL, perform near ceiling on the original five test suites and are not affected at all by intervening content. 

\subsection{Effects of model pre-processing}
\label{sec:bpe}

The GPT-2 models trained and evaluated in this paper use a sub-word vocabulary learned by byte-pair encoding \citep[BPE; ][]{sennrich-etal-2016-neural} to represent their inputs, while all other models represent and compute over word-level inputs. This byte-pair encoding was taken from the pre-trained GPT-2 model trained on a much larger corpus.
%GPT-2's extremely high performance in SG score 
The results reported for these models thus conflate a choice of model class (a deep Transformer architecture) and preprocessing standard (sub-word tokenization computed on a larger corpus). Some preliminary work suggests that sub-word tokenization is indeed responsible for much of the larger GPT-2 models' success: we find that GPT-2 models trained on word-level representations of \blg and \bmd achieve good perplexity measures, but degrade sharply in SG score. 

Peculiarities of the GPT-2 training regime may be responsible for its particularly bad performance on the smaller corpora. Its sub-word vocabulary was held constant across training corpora, meaning that the model vocabulary size also remained constant across corpora, unlike the other models tested.
%Unlike other model architectures, GPT-2's sub-word vocabulary was not changed as it was trained on corpora of different sizes.
The poor performance of GPT-2 models trained on smaller corpora may thus be due to overparameterization, and not due to fundamental problems with the model architecture at small data scales.
%\todo{Recall from \Cref{sec:results-SG-perplexity} that the larger controlled GPT-2 models perform exceptionally well on both SG score and perplexity, and the smaller ones perform exceptionally poorly. This suggests an interaction between model class and data scale below some minimum data threshold for some models. One potential explanation for this trend is that the GPT-2 models are trained on a representation of BLLIP using sub-words from the pre-trained GPT-2 model, tacitly importing information from a larger training dataset. The asymmetry between BLLIP and GPT-2 vocabularies may unfairly benefit the \blg and \bmd models, while harming the \bsm and \bxs models.}
We leave a thorough investigation of the role of sub-word tokenization to future work.

\section{Discussion}

%We presented a scaled-up and controlled pipeline for evaluating the syntactic generalization abilities of neural language models. We collected 34 syntactic generalization test suites and used them to evaluate 16 model-class$\times$training-corpus combinations, allowing for the first controlled comparison between broad classes of neural language models for these tests.

This work addresses multiple open questions about syntactic evaluations and their relationship to other language model assessments. Our results dissociate model perplexity and performance in syntactic generalization tests, suggesting that the two metrics capture complementary features of language model knowledge. In a controlled evaluation of different model classes and datasets, we find model architecture plays a more important role than training data scale in yielding correct syntactic generalizations.
% although large-data models are able to %achieve both lower perplexity and higher SG scores.
%Comparing our controlled models across syntactic circuits 
Our circuit-level analysis reveals consistent failure on Licensing but inconsistent behavior on other circuits, suggesting that different syntactic circuits make use of different underlying processing capacities. %variation in model performance is due to improvements on a smaller number of critical circuits, which test models on their ability to deal with complex structural analyses of their input sentences. 
In addition to the insight these results provide about neural NLP systems, they also bear on questions central to cognitive science and linguistics, putting lower bounds on what syntactic knowledge can be acquired from string input alone.

Targeted syntactic evaluation is just one in a series of complementary methods being developed to assess the learning outcomes of neural language processing models. Other methods include classifying sentences as grammatical or ungrammatical \citep{warstadt2018neural}, decoding syntactic features from a model's internal state \citep{belinkov2017neural, giulianelli2018under}, or transfer learning to a strictly syntactic task such as parsing or POS tagging \citep{hewitt2019structural}. As each task brings an explicit set of assumptions, %---about e.g.\ the structure of natural language, or the adaptation of a syntactic formalism as a `gold standard'---
complementary assessment methods can collectively provide greater insight into models' learning outcomes. %The psycholinguistic approach to neural LM assessment is valuable because (1) it directly measures the extent to which a models' learned distribution conforms to real-time processing behavior, and (2) has already been used to assess neural LM learning of a wide range of structural properties  \citep{Linzen:et-al:2016,Kuncoro:et-al:2017,Lau:et-al:2017,Futrell:et-al:2018,Gulordava:et-al:2018,Marvin:Linzen:2018,Wilcox:et-al:2018,Futrell:et-al:2019,Wilcox:et-al:2019}. 
Although this paper, together with \citet{warstadt2019blimp}, report what is to our knowledge the largest-scale targeted syntactic evaluations to date, we emphasize that they are only first steps
toward a comprehensive understanding of the syntactic capabilities of contemporary language models. This understanding will be further advanced by new targeted-evaluation test suites covering a still wider variety of syntactic phenomena, additional trained models with more varied hyperparameters and randomization seeds, and new architectural innovations.  Humans develop extraordinary grammatical capabilities through exposure to natural linguistic input.  It remains to be seen to just what extent contemporary artificial systems do the same.

%\pengcomment{the components that lead to better and more robust syntactic generalization of neural language models. }

\section*{Acknowledgments}

The authors would like to thank the anonymous reviewers and Samuel R. Bowman for their feedback, Miguel Ballesteros for advice and technical guidance, and Tristan Thrush for technical assistance. J.H.~is supported by the NIH under award number T32NS105587 and an NSF Graduate Research Fellowship. J.G.~is supported by an Open Philanthropy AI Fellowship. R.P.L.~gratefully acknowledges support from the MIT-IBM Watson AI Lab, a Google Faculty Research Award, and a Newton Brain Science Award.

\bibliography{acl2020} %,anthology}
\bibliographystyle{acl_natbib}

\appendix

\section{Syntactic coverage of test suites} \label{sec:coverage}

\input{syntactic_coverage.tex}

\section{Description of test suites} \label{sec:summary-test-suites}

\input{test_suites_appendix.tex}

% \section{Circuit--circuit correlations} \label{sec:circuit-correlations}
% \input{correlation_appendix.tex}

\end{document}

%% file: syntactic_coverage.tex
\begin{table*}[p]
    \centering
    % \small
    \resizebox{\linewidth}{!}{
    \begin{tabular}{llc} \toprule
        \textsc{Chapter 1: Generative Grammar} & Lexical gender & \\
        & Number & \checkmark \\
        & Person & \\
        & Case & \\ \midrule
        \textsc{Chapter 2: Parts of Speech} & Parts of Speech & \checkmark \\
        & Plurality & \checkmark \\
        & Count vs. Mass Nouns & \\
        & Argument Structure of Verbs & \checkmark \\ \midrule
        \textsc{Chapter 3: Constituency, Trees, Rules} & Constituency Tests & \\
        & Hierarchical Structure & \checkmark \\ \midrule
        \textsc{Chapter 4: Structural Relations} & c-command & \checkmark \\
        & Government & \\ \midrule
        \textsc{Chapter 5: Binding Theory} & $R$-expression vs. Pronominals & \\
        & Anaphoric expressions and their antecedents & \checkmark \\
        & Co-reference and co-indexation & \\
        & Binding Principles ($A, B, C$) & \checkmark \\
        & Locality Constraints & \checkmark \\ \midrule
        \textsc{Chapter 6: X-Bar Theory} & One Replacement & \\
        & Do-so Replacement & \\ \midrule
        \textsc{Chapter 7: Extending X-Bar Theory} & Fundamental Phrase Types of DP/CP/TP & \\
        \textsc{to Functional Categories} & Genitives: of-genitives and 's genitives & \\
        & Subjects and Predicates & \\
        & Clausal Embedding & \checkmark \\ 
        & Clausal & \\
        & Tense/Finiteness and its restrictions & \\
        & Yes/No Questions & \\
        & Subject-Auxilliary Inversion & \\ \midrule
        \textsc{Chapter 8: Constraining X-Bar Theory:} & Thematic Relations & \checkmark \\
        \textsc{The Lexicon} & Internal Theta role vs. External Theta Roles & \\
        & Expletive Pronouns and Expletive Insertion & \\
        & Extended Projection Principle & \\ \midrule
        \textsc{Chapter 9: Head-to-Head Movement} & V $\rightarrow$ T Movement & \\
        & T $\rightarrow$ C movement & \checkmark \\
        & Do-Support & \\ \midrule
        \textsc{Chapter 10: DP Movement} & Passive Constructions & \checkmark \\
        & DP-Raising & \\ \midrule
        \textsc{Chapter 11: Wh-Movement} & Wh-Movement & \checkmark \\
        & Structural Constraints on Wh-Movement (Island Constraints) & \checkmark \\
        & Wh in-Situ and Echo Questions \\ \midrule
        \textsc{Chapter 12: A Unified Theory} & Universal Quantifiers vs. Existential Quantifiers & \\
        \textsc{of Movement} & Quantificational Scope and Quantifier Raising & \\ \midrule
        \textsc{Chapter 13: Extended VPs} & Light Verbs & \\
        & Object Shift (and end weight) & \\
        & Ellipsis & \\
        & Pseudogapping & \\ \midrule
        \textsc{Chapter 14: Raising Control and} & Control, Subject-to-Subject and Subject-to-Object Raising (ECM) \\ 
        \textsc{Empty Categories} & & \\ \midrule
        \textsc{Chapter 15: Advanced Topics in} & Binding Principle $A$ and $B$ & \checkmark  \\ 
        \textsc{Binding Theory} & & \\ \bottomrule
    \end{tabular}
    }
    \caption{Test suite coverage of syntactic phenomena presented in \citet{carnie2012syntax}.}
    \label{tab:coverage}
\end{table*}

In order to assess the coverage of our syntactic tests, we manually inspected the ``Ideas, Rules and Constraints introduced in this Chapter'' section for each chapter in \citet{carnie2012syntax}, a standard introductory syntax textbook. We included entries from these sections which are theory-neutral and refer to observable linguistic data. For example, we do not include \textit{affix lowering} (Chapter 7) or \textit{theta criterion} (Chapter 8) because these phenomena presuppose a commitment to one particular syntactic analysis.

We found that our tests covered 16 of the 47 phenomena presented ($\sim$34\%). Of the 15 chapters surveyed, our tests assessed phenomena in 11 ($\sim$73\%). We did not assess coverage from the last two chapters of the book, which explore alternative syntactic formalisms. The outcome of our manual inspection is given in \Cref{tab:coverage}.

A \checkmark indicates that some aspect of that phenomena was tested in one or more of our suites. \checkmark does not necessarily mean that the test suite was designed explicitly for the purpose of testing that phenomena, but merely that the phenomena was implicated in model success. For example, we place a \checkmark next to \textit{Parts of Speech} because differentiation between verbs and nouns is necessary for models to succeed in the \textit{Cleft Structure} tests.

%% file: test_suites_appendix.tex
\setlength{\Extopsep}{.25\baselineskip}
\setlength{\Exlabelsep}{0.75cm} 
\setlength{\Exindent}{0cm} 

\renewcommand{\theExNo}{\Alph{ExNo}}%modified in version 4.0
\renewcommand{\Exarabic}{\Alph}%added in version 4.0

\newcommand{\Pcond}[1]{\ensuremath{P_{\text{\ref{#1}}}}}
\newcommand{\Scond}[1]{\ensuremath{S_{\text{\ref{#1}}}}}

% \setcounter{section}{1}

% \section*{Appendix A: Summary of test suites}
% \label{sec:summarry-test-suites}

In this work we have assembled a large number of test suites inspired
by the methodology of experimental sentence-processing and
psycholinguistic research.  Each test suite contains a number of
\key{items}, and each item appears in several \key{conditions}: across
conditions, a given item will differ only according to a controlled
manipulation designed to target a particular feature of
grammatical knowledge. For each suite we define a \key{success criterion}, which stipulates inequalities among conditional probabilities of sentence substrings.

In the main paper, a model's accuracy for a
test suite is computed as the percentage of the test suite's items for
which it satisfies the criterion. In this appendix, we briefly describe each test suite and the criterion used to
determine whether a given model succeeds on each item of the test
suite.

\subsection{Notation}
\label{sec:notation}

\subsubsection{Sentence status}
\label{sec:sentence-status}

Following and building on linguistic traditions, we annotate
examples as follows.  Examples marked with a *
violate a well-established grammatical constraint, and are
ungrammatical.  Examples marked with ?\ or ??\ are not necessarily
ungrammatical, but are marginal: for example, they may require an
unusual interpretation of a word in order for the sentence to be
grammatical. (More ?'s is roughly intended to indicate more severe
marginality).  Examples marked with !\ are not ungrammatical, but
induce severe processing difficulty that is measurable in real-time
human sentence processing.  For all test suites, we include references
to established literature on the relevant grammatical and/or
sentence-processing phenomena.

\subsubsection{Success criteria}
\label{sec:notat-regard-pred}

Criteria involve inequalities among conditional probabilities
of sentence substrings given the complete sentence context preceding
the substring.  In describing criteria, we use $P(\cdot)$ for raw
probabilities and $S(\cdot)$ for surprisals (negative
log-probabilities), and leave the conditioning on preceding context
implicit.  For concision, we use subscripts on $P$ and $S$ to indicate
the variant of the sentence within the test suite that we are
referring to.  In the first described test suite, \key{center
  embedding} \ref{sec:center-embedding}, we show the criterion in both
concise and fully spelled-out forms, to help clarify the conventions
we are using in the concise form.  All items within a given test suite share the same criterion for success.

We provide chance accuracy on the assumption that the order of probabilities among conditions for a given item is random.  In some cases, exactly determining chance accuracy may require further assumptions about the distribution of these probabilities; in this case we provide an upper bound on chance accuracy.

\subsection{Center embedding}
\label{sec:center-embedding}

Center embedding, the ability to embed a phrase in the middle of
another phrase of the same type, is a hallmark feature of natural
language syntax.  Center-embedding creates \key{nested syntactic
  dependencies}, which could pose a challenge for some language
models.  To succeed in generating expectations about how sentences
will continue in the context of multiple center embedding, a model
must maintain a representation not only of what words appear in the
preceding context but also of the order of those words, and must
predict that upcoming words occur in the appropriate order.  In this
test suite we use verb transitivity and subject--verb plausibility to
test model capabilities in this respect.  For example,
\ref{ex:center-embedding-correct} below is a correct center-embedding,
but \ref{ex:center-embedding-incorrect} is not:

\setcounter{ExNo}{0}

\ex. The painting$_{\text{N}_1}$ that the artist$_{\text{N}_2}$
painted$_{\text{V}_2}$ deteriorated$_{\text{V}_1}$. [correct] \label{ex:center-embedding-correct}

\ex. ??The painting$_{\text{N}_1}$ that the artist$_{\text{N}_2}$
deteriorated$_{\text{V}_1}$ painted$_{\text{V}_2}$. [incorrect] \label{ex:center-embedding-incorrect}

\noindent
Here, N$_i$ and $V_i$ correspond to matched subject--verb pairs.

% \ex.
% \a. The painting$_{\text{N}_1}$ that the artist$_{\text{N}_2}$
% painted$_{\text{V}_2}$ deteriorated$_{\text{V}_1}$. [correct] \label{ex:center-embedding-correct}
% \b. \#The painting$_{\text{N}_1}$ that the artist$_{\text{N}_2}$
% deteriorated$_{\text{V}_1}$ painted$_{\text{V}_2}$. [incorrect] \label{ex:center-embedding-incorrect}
% \z.

In the \textsc{with-modifier} version of the test suite, we postmodify
N$_2$ with a relative clause to increase the linear distance over
which the nested dependencies must be tracked, potentially leading to
a harder test suite:

\setcounter{ExNo}{0}

\ex. The painting$_{\text{N}_1}$ that the artist$_{\text{N}_2}$ who
lived long ago
painted$_{\text{V}_2}$ deteriorated$_{\text{V}_1}$. [correct] 

\ex. \#The painting$_{\text{N}_1}$ that the artist$_{\text{N}_2}$ who
lived long ago
deteriorated$_{\text{V}_1}$ painted$_{\text{V}_2}$. [incorrect]

\paragraph{Criterion} The probability of the verb sequence in the
correct variant should be higher than the probability of the verb
sequence in the incorrect variant:
\begin{align*}
\Pcond{ex:center-embedding-correct}(\text{V}_2 \text{V}_1) >
\Pcond{ex:center-embedding-incorrect}(\text{V}_1 \text{V}_2)  
\end{align*}

\noindent In
full form, this criterion for the example item in the no-modifier
version of this  test suite would be:
{\small{\begin{align*}
&  P(\text{painted deteriorated}|\text{The painting that the artist}) >
  \\
&  P(\text{deteriorated painted}|\text{The painting that the artist})
\end{align*}}}
\noindent Chance performance on these center-embedding test suites would be
50\%.

\paragraph{References} \textcite{miller-chomsky:1963};\textcite{wilcox-etal:2019-hierarchical-representation}

\subsection{Pseudo-clefting}
\label{sec:np-vp-clefting}

\setcounter{ExNo}{0}

The pseudo-cleft construction involves (i) an extraction of a
\key{targeted constituent} from a sentence and (ii) a constituent that
provides the semantic contents of the targeted constituent and must
match it in syntactic category, where (i) and (ii) are linked by the
copula.  The pseudo-cleft construction can target both NPs and VPs; in
the latter case, the VP of the free relative becomes an inflected form
of \emph{do}.  This means that a free relative subject plus the copula can set
up a requirement for the syntactic category that comes next.  If the
free relative clause has a \emph{do} VP without a direct object, then
the main-clause postcopular predicate can be a VP (\ref{ex:clefting-vp-match} below). Otherwise, the postcopular predicate
must be an NP (\ref{ex:clefting-np-match} below):

\newcommand{\pseudocleftedVP}{\ensuremath{\text{VP}}}
\newcommand{\pseudocleftedNP}{\ensuremath{\text{NP}}}

\ex. What the worker did  was $\overbrace{\text{board the
    plane}}^{\pseudocleftedVP}$.  \label{ex:clefting-vp-match}

\ex. ?What the worker did  was $\overbrace{\text{the plane}}^{\pseudocleftedNP}$.  \label{ex:clefting-vp-mismatch}

\ex. What  the worker repaired was $\overbrace{\text{the
    plane}}^{\pseudocleftedNP}$. \label{ex:clefting-np-match}

\ex.  *What  the worker repaired was $\overbrace{\text{board the
    plane}}^{\pseudocleftedVP}$. \label{ex:clefting-np-mismatch}

\paragraph{Criterion} The postcopular predicate should be
more surprising when its syntactic category mismatches the cleft, averaging across VP and NP postcopular predicates:

\begin{align*}
\Scond{ex:clefting-np-mismatch}(\pseudocleftedVP) +
\Scond{ex:clefting-vp-mismatch}(\pseudocleftedNP) >
\Scond{ex:clefting-np-match}(\pseudocleftedNP) +
\Scond{ex:clefting-vp-match}(\pseudocleftedVP)  
\end{align*}

\noindent Chance is 50\%.  A more
stringent criterion would be to apply this requirement separately for each of NP and VP postcopular predicates: 
\begin{align*}
\Scond{ex:clefting-np-mismatch}(\pseudocleftedVP)
  >\Scond{ex:clefting-vp-match}(\pseudocleftedVP) \wedge
\Scond{ex:clefting-vp-mismatch}(\pseudocleftedNP) >
\Scond{ex:clefting-np-match}(\pseudocleftedNP)
\end{align*}
However, it is often possible to use an NP postcopular predicate with
a \emph{do} cleft through semantic coercion (e.g., in
\ref{ex:clefting-vp-mismatch} ``did'' can be interpreted as ``fixed''
or ``was responsible for''), so we felt that this latter criterion
might be too stringent.

\paragraph{References} \textcite{higgins:1973-pseudo-cleft}

\subsection{Filler--gap dependencies}
\label{sec:fill-gap-depend}

Consider the following sentence, in which all arguments and adjuncts
appear ``in situ'' (in the syntactic position at which they are
normally interpreted semantically):
\begin{quote}
  I know that our uncle grabbed the food in front of the guests at the
  holiday party.
\end{quote}
A \key{filler--gap dependency} can be created by \key{extracting} any
of a number of elements from the subordinate clause, including
\emph{our uncle} (subject extraction), \emph{the food} (object
extraction) or \emph{the guests} (extraction from a prepositional
phrase).  These possibilities serve as the basis for several test
suites on filler--gap dependencies.

\paragraph{References}
\textcite{ross:1967,crain-fodor:1985how-can-grammars,stowe:1986parsing,wilcox-etal:2018-what-do-rnns,chowdhury-zamparelli:2018-rnn,chaves:2020-what-dont}

\subsubsection{Subject extractions}
\label{sec:subject-extractions}

\setcounter{ExNo}{0}

\ex.  I know that $\overbrace{\text{our uncle}}^{\alpha}$ grabbed the food in front of the guests at the
holiday party. [\textsc{that}, \textsc{no gap}] \label{ex:fgd-subj-that-nogap}

\ex.  *I know who $\overbrace{\text{our uncle}}^{\alpha}$  grabbed the food in front of the guests at the
holiday party. [\textsc{wh}, \textsc{no gap}] \label{ex:fgd-subj-wh-nogap}

\ex.  *I know that $\overbrace{\text{grabbed}}^{\beta}$ the food in front of the guests at the
holiday party. [\textsc{that}, \textsc{gap}] \label{ex:fgd-subj-that-gap}

\ex.  I know who  $\overbrace{\text{grabbed}}^{\beta}$  the food in front of the guests at the
holiday party. [\textsc{wh}, \textsc{gap}] \label{ex:fgd-subj-wh-gap}

\paragraph{Criterion} We require that a model successfully pass a
two-part criterion for each item: the \emph{wh-}filler should
make the unextracted subject $\alpha$ more surprising in the \textsc{no-gap}
conditions  and should make the post-gap material $\beta$ less
surprising in the \textsc{gap} conditions:
\begin{align*}
  \Scond{ex:fgd-subj-wh-nogap}(\alpha) >
  \Scond{ex:fgd-subj-that-nogap}(\alpha) \wedge   \Scond{ex:fgd-subj-that-gap}(\beta) >
  \Scond{ex:fgd-subj-wh-gap}(\beta)
\end{align*}
Chance is 25\%.

\subsubsection{Object extractions}
\label{sec:object-extractions}

\setcounter{ExNo}{0}

The logic of this test suite is the same as that for subject
extraction above.  Note that we use obligatorily transitive embedded verbs, so
that omitting a direct object should be highly surprising when there
is no filler, as in \ref{ex:fgd-obj-that-gap}.

\ex.  I know that our uncle grabbed $\overbrace{\text{the food}}^{\alpha}$ in front of the guests at
the holiday party. [\textsc{that}, \textsc{no
  gap}] \label{ex:fgd-obj-that-nogap}

\ex.  *I know what our uncle grabbed $\overbrace{\text{the food}}^{\alpha}$ in front of the guests at
the holiday party. [\textsc{wh}, \textsc{no
  gap}] \label{ex:fgd-obj-wh-nogap}

\ex.  ??I know that our uncle grabbed $\overbrace{\text{in front of}}^{\beta}$ the guests at
the holiday party. [\textsc{that}, \textsc{gap}] \label{ex:fgd-obj-that-gap}

\ex.  I know what our uncle grabbed $\overbrace{\text{in front of}}^{\beta}$ in front of the guests at
the holiday party. [\textsc{wh}, \textsc{gap}] \label{ex:fgd-obj-wh-gap}

\paragraph{Criterion}
\begin{align*}
  \Scond{ex:fgd-obj-wh-nogap}(\alpha) >
  \Scond{ex:fgd-obj-that-nogap}(\alpha) \wedge   \Scond{ex:fgd-obj-that-gap}(\beta) >
  \Scond{ex:fgd-obj-wh-gap}(\beta)
\end{align*}

\subsubsection{Extraction from prepositional phrases}
\label{sec:extractions-from-pp}

\setcounter{ExNo}{0}

The logic of this test suite is the same as that for subject
and object extractions above.  

\ex.  I know that our uncle grabbed the food in front of $\overbrace{\text{the guests}}^{\alpha}$ at
the holiday party. [\textsc{that}, \textsc{no
  gap}] \label{ex:fgd-pp-that-nogap}

\ex.  *I know who our uncle grabbed the food in front of $\overbrace{\text{the guests}}^{\alpha}$  at
the holiday party. [\textsc{wh}, \textsc{no
  gap}] \label{ex:fgd-pp-wh-nogap}

\ex.  *I know that our uncle grabbed the food in front of $\overbrace{\text{at
the holiday party}}^{\beta}$. [\textsc{that}, \textsc{gap}] \label{ex:fgd-pp-that-gap}

\ex.  I know who our uncle grabbed the food in front of $\overbrace{\text{at
the holiday party}}^{\beta}$. [\textsc{wh}, \textsc{gap}] \label{ex:fgd-pp-wh-gap}

\paragraph{Criterion}
\begin{align*}
  \Scond{ex:fgd-pp-wh-nogap}(\alpha) >
  \Scond{ex:fgd-pp-that-nogap}(\alpha) \wedge   \Scond{ex:fgd-pp-that-gap}(\beta) >
  \Scond{ex:fgd-pp-wh-gap}(\beta)
\end{align*}

\subsubsection{Tests for unboundedness}
\label{sec:tests-unboundedness}

Filler--gap dependencies are ``unbounded'' in the sense that there is
no limit to how many clausal levels above the gap the filler can be
extracted.  This serves as  the basis for harder versions of the
object-extracted test suites, involving three or four levels of clausal
embedding.  Example [\textsc{that, no gap}] sentences are given below:

\begin{quote}
  I know that our mother said her friend remarked that the park
  attendant reported your friend threw the plastic into the trash
  can. [3 levels of embedding]
\end{quote}

\begin{quote}
I know that our mother said her friend remarked that the park
attendant reported the cop thinks your friend threw the plastic into
the trash can. [4 levels of embedding]
\end{quote}

These base sentences give rise to 4-condition test suites using the
same manipulations as for the basic object-extraction test suite
(Section~\ref{sec:object-extractions}), and the criterion for success is the same.

\subsection{Main-verb/reduced-relative garden-path disambiguation}
\label{sec:main-verbr-relat}

\newcommand{\disambiguator}{\ensuremath{\text{V}^*}\xspace}

\setcounter{ExNo}{0}

This is one of the best-studied instances of syntactic garden-pathing
in the psycholinguistics literature.  An example 4-condition  item is
given below:

\ex. !The child kicked in the chaos
$\overbrace{\text{found}}^{\disambiguator}$ her way back
home. [\textsc{reduced}, \textsc{ambig}] \label{ex:mv-rr-reduced-ambig}

\ex. The child who was kicked in the chaos
$\overbrace{\text{found}}^{\disambiguator}$ her way back home.  \label{ex:mv-rr-unreduced-ambig}

\ex. The child forgotten in the chaos
$\overbrace{\text{found}}^{\disambiguator}$ her way back home. \label{ex:mv-rr-reduced-unambig}

\ex. The child who was forgotten in the chaos
$\overbrace{\text{found}}^{\disambiguator}$ her way back home. \label{ex:mv-rr-unreduced-unambig}

\paragraph{Criterion} Relative to the [\textsc{reduced},
\textsc{ambig}] condition, not reducing the relative clause should
make \disambiguator less surprising, as should changing the
participial verb to one that is the same form as a simple past-tense
verb.  Additionally, the effect of not reducing the relative clause on
\disambiguator surprisal should be smaller for unambiguous participial
verbs than for participial verbs:
\begin{align*}
  \Scond{ex:mv-rr-reduced-ambig}(\disambiguator) > \Scond{ex:mv-rr-unreduced-ambig}(\disambiguator)
  \wedge   \Scond{ex:mv-rr-reduced-ambig}(\disambiguator) > \Scond{ex:mv-rr-reduced-unambig}(\disambiguator)
  \wedge \\
  \Scond{ex:mv-rr-reduced-ambig}(\disambiguator) -
  \Scond{ex:mv-rr-unreduced-ambig}(\disambiguator) > \Scond{ex:mv-rr-reduced-unambig}(\disambiguator) - \Scond{ex:mv-rr-unreduced-unambig}(\disambiguator)
\end{align*}
Chance is somewhere below 25\%. 
% \textbf{[TBD: does true chance depends on
%   assumptions about surprisal distributions?}

\paragraph{References} \textcite{bever:1970,ferreira-clifton:1986,trueswell-etal:1994,van-schijndel-linzen:2018cogsci,futrell-etal:2019-neural-language-models}

\subsection{Negative Polarity Licensing}
\label{sec:negat-polar-licens}

\newcommand{\NPI}{\ensuremath{\text{NPI}}\xspace}

\setcounter{ExNo}{0}

The words \emph{any} and \emph{ever}, in their most common uses, are
``negative polarity items'' (NPIs): they can only be used in an
appropriate syntactic-semantic environment---to a first approximation,
in the scope of negation.  For example, the determiner \emph{no} can
license NPIs, but its NP has to structurally command the NPI.  Below,
\ref{ex:npi-src-neg-pos} and \ref{ex:npi-src-neg-neg} are acceptable,
because \emph{no} is the determiner for the subject noun
\emph{managers}.  There is no negation in \ref{ex:npi-src-pos-pos} so
the NPI is unlicensed and the sentence is unacceptable; crucially,
however, \ref{ex:npi-src-pos-neg} is unacceptable despite the presence
of \emph{no} earlier in the sentence, because \emph{no} is embedded
inside a modifier of the main-clause subject and thus does not command
the NPI.  

\ex. No managers that respected the guard have had
$\overbrace{\text{any}}^{\text{NPI}}$ luck. [\textsc{+neg,--distractor}]\label{ex:npi-src-neg-pos}

\ex. *The managers that respected no guard have had
$\overbrace{\text{any}}^{\text{NPI}}$ luck. [\textsc{--neg,+distractor}]\label{ex:npi-src-pos-neg}

\ex. *The managers that respected the guard have had
$\overbrace{\text{any}}^{\text{NPI}}$ luck. [\textsc{--neg,--distractor}]\label{ex:npi-src-pos-pos}

\ex. No managers that respected no guard have had
$\overbrace{\text{any}}^{\text{NPI}}$ luck. [\textsc{+neg,+distractor}]\label{ex:npi-src-neg-neg}

In the above test suite, the ``distractor'' position for \emph{no} is
inside a subject-extracted relative clause modifying the main-clause
subject.  We also used a variant test suite in which these relative
clauses are object-extracted:

\setcounter{ExNo}{0}

\ex. No managers that  the guard respected have had
$\overbrace{\text{any}}^{\text{NPI}}$ luck. [\textsc{+neg,--distractor}]\label{ex:npi-any-src-neg-pos}

\ex. *The managers that  no guard respected have had
$\overbrace{\text{any}}^{\text{NPI}}$ luck. [\textsc{--neg,+distractor}]\label{ex:npi-any-src-pos-neg}

\ex. *The managers that  the guard respected have had
$\overbrace{\text{any}}^{\text{NPI}}$ luck. [\textsc{--neg,--distractor}]\label{ex:npi-any-src-pos-pos}

\ex. No managers that no guard respected have had
$\overbrace{\text{any}}^{\text{NPI}}$ luck. [\textsc{+neg,+distractor}]\label{ex:npi-any-src-neg-neg}

The above two test suites use \emph{any} as the NPI; we also use test
suites with \emph{ever} as the NPI.  Subject-extracted relative clause example:

\setcounter{ExNo}{0}

\ex. No managers that respected the guard have
$\overbrace{\text{ever}}^{\text{NPI}}$ gotten old. [\textsc{+neg,--distractor}]\label{ex:npi-ever-src-neg-pos}

\ex. *The managers that respected no guard have 
$\overbrace{\text{ever}}^{\text{NPI}}$ gotten old. [\textsc{--neg,+distractor}]\label{ex:npi-ever-src-pos-neg}

\ex. *The managers that respected the guard have
$\overbrace{\text{ever}}^{\text{NPI}}$ gotten old. [\textsc{--neg,--distractor}]\label{ex:npi-ever-src-pos-pos}

\ex. No managers that respected no guard have 
$\overbrace{\text{ever}}^{\text{NPI}}$ gotten old. [\textsc{+neg,+distractor}]\label{ex:npi-ever-src-neg-neg}

Object-extracted relative clause example:

\setcounter{ExNo}{0}

\ex. No managers that  the guard respected have
$\overbrace{\text{ever}}^{\text{NPI}}$ gotten old. [\textsc{+neg,--distractor}]\label{ex:npi-ever-orc-neg-pos}

\ex. *The managers that  no guard respected have 
$\overbrace{\text{ever}}^{\text{NPI}}$ gotten old. [\textsc{--neg,+distractor}]\label{ex:npi-ever-orc-pos-neg}

\ex. *The managers that  the guard respected have 
$\overbrace{\text{ever}}^{\text{NPI}}$ gotten old. [\textsc{--neg,--distractor}]\label{ex:npi-ever-orc-pos-pos}

\ex. No managers that no guard respected have 
$\overbrace{\text{ever}}^{\text{NPI}}$ gotten old. [\textsc{+neg,+distractor}]\label{ex:npi-ever-orc-neg-neg}

\paragraph{Criterion} Changing the main-clause subject's determiner
from \emph{The} to \emph{No} should increase the probability of the
NPI where it appears, regardless of whether there is a distractor
\emph{no} in the subject-modifying relative clause.  Furthermore, when
there is exactly one \emph{no} in the sentence, the NPI should be
higher-probability when it is in a licensing position rather than in a
distractor position:
\begin{align*}
&  \Pcond{ex:npi-ever-orc-neg-pos}(\NPI) >
  \Pcond{ex:npi-ever-orc-pos-pos}(\NPI) \wedge   \Pcond{ex:npi-ever-orc-neg-neg}(\NPI) >
  \Pcond{ex:npi-ever-orc-pos-neg}(\NPI) \wedge  \\
&  \Pcond{ex:npi-ever-orc-neg-pos}(\NPI) >
  \Pcond{ex:npi-ever-orc-pos-neg}(\NPI)
\end{align*}
%
%f <- function(x) {
%  return((which.max("a"==x) < which.max("c"==x)) && 
%         (which.max("d"==x) < which.max("b"==x)) && 
%         (which.max("a"==x) < which.max("b"==x)))
%}
%library(combinat)
%mean(sapply(permn(letters[1:4]),f))
%
Chance is $\frac{5}{32}$.
%\textbf{[TBD: does true chance depends on
%   assumptions about surprisal distributions?}

\paragraph{References}
\textcite{ladusaw:1979polarity,vasishth-etal:2008,giannakidou:2011negative,Marvin:Linzen:2018,Futrell:et-al:2018}

\subsection{NP/Z garden-path  ambiguity}
\label{sec:npz-garden-path}

\setcounter{ExNo}{0}

This is another well-studied syntactic garden-pathing
configuration. In \ref{ex:npz-trans-no-comma} below, the NP \emph{the
  waters} introduces a local syntactic ambiguity: it could be (1) the
direct object of \emph{crossed}, in which case the sentence-initial
subordinate clause has not yet ended, or (2) the subject of the main
clause, in which case \emph{crossed} is used intransitively and is the
last word of the sentence-initial subordinate clause. (This was dubbed
``NP/Z'' by \textcite{sturt-etal:1999} because the subordinate-clause
verb might have either an NP object or a Z(ero), i.e.\ null, object.)
The next word, \emph{remained}, is only compatible with (2); the
ruling out of (1) generally yields increased processing difficulty for
human comprehenders.  Marking the end of the subordinate clause with a
comma, as in \ref{ex:npz-trans-comma}, makes the sentence easier at
\disambiguator, as does an obligatorily intransitive
subordinate-clause verb, as in \ref{ex:npz-intrans-no-comma}.

\ex. !As the ship crossed the waters
$\overbrace{\text{remained}}^{\disambiguator}$ blue and
calm. [\textsc{trans},\textsc{no comma}] \label{ex:npz-trans-no-comma}

\ex. As the ship crossed, the waters
$\overbrace{\text{remained}}^{\disambiguator}$ blue and
calm. [\textsc{trans},\textsc{comma}] \label{ex:npz-trans-comma}

\ex. As the ship drifted the waters
$\overbrace{\text{remained}}^{\disambiguator}$ blue and
calm. [\textsc{intrans},\textsc{no comma}] \label{ex:npz-intrans-no-comma}

\ex. As the ship drifted, the waters
$\overbrace{\text{remained}}^{\disambiguator}$ blue and
calm. [\textsc{intrans},\textsc{comma}] \label{ex:npz-intrans-comma}

\paragraph{Criterion} Similar to the main-verb/reduced-relative
garden-pathing ambiguity, a model must pass a three-part criterion.  Relative to \ref{ex:npz-trans-no-comma}, either marking the
subordinate-clause end with a comma or using an obligatorily
intransitive verb in the subordinate clause should reduce the
surprisal of \disambiguator.  Furthermore, the surprisal-reduction
effect of the comma should be smaller when the subordinate-clause verb
is intransitive than when it is transitive:
% \begin{align*}
%   \Scond{ex:npz-trans-no-comma} >  \Scond{ex:npz-trans-comma} \wedge
%     \Scond{ex:npz-trans-no-comma} >  \Scond{ex:npz-intrans-no-comma}
%   \wedge\\
%   \Scond{ex:npz-trans-no-comma} >  \Scond{ex:npz-trans-comma} >   \Scond{ex:npz-intrans-no-comma} -  \Scond{ex:npz-intrans-comma}
% \end{align*}
\begin{align*}
  \Scond{ex:npz-trans-no-comma}(\disambiguator) >
  \Scond{ex:npz-trans-comma}(\disambiguator) \wedge
  \Scond{ex:npz-trans-no-comma}(\disambiguator) >
  \Scond{ex:npz-intrans-no-comma}(\disambiguator) \wedge \\
   \Scond{ex:npz-trans-no-comma}(\disambiguator) -
  \Scond{ex:npz-trans-comma}(\disambiguator) >  \Scond{ex:npz-intrans-no-comma}(\disambiguator) -   \Scond{ex:npz-intrans-comma}(\disambiguator)
\end{align*}

We also use an NP/Z test suite where the second means of
disambiguation is not changing the subordinate-clause verb to an
intransitive, but rather giving the transitive subordinate-clause verb
an overt direct object.  For the above example item,  the first two
conditions are the same and the other two conditions would be:

\setcounter{ExNo}{2}

\ex. As the ship crossed the sea the waters
$\overbrace{\text{remained}}^{\disambiguator}$ blue and
calm. 

\ex. As the ship crossed the sea, the waters
$\overbrace{\text{remained}}^{\disambiguator}$ blue and
calm.

\noindent
The success criterion remains the same.

Finally, we create harder versions of both the above test suites by
adding a postmodifier to the main-clause subject (in the above
example, \emph{the waters} becomes \emph{the waters of the Atlantic
  Ocean}).

\paragraph{References} \textcite{frazier-rayner:1982,mitchell:1987,pickering-traxler:1998,sturt-etal:1999,staub:2007}

\subsection{Subject--verb number agreement}
\label{sec:subject-verb-number}

\setcounter{ExNo}{0}

This task tests a language model for how well it predicts the number
marking on English finite present-tense verbs (whether it should be
the third-person \emph{singular} form, or the
non-third-person-singular form, generally referred to as the
\emph{plural} form for simplicity, although technically this is the
form for first- and second-person singular as well).  In controlled,
targeted versions of this test, multiple NP precede the verb: the
verb's actual subject, as well as a \key{distractor} NP with number
that is different from that of the subject.  A successful language
model should place higher probability on the verbform matching that of
the subject, not the distractor.  We have three versions of this test
suite: one where the distractor is in a prepositional phrase
postmodifier of the subject:

\newcommand{\Vsg}{\ensuremath{\text{V}_{\text{sg}}\xspace}}
\newcommand{\Vpl}{\ensuremath{\text{V}_{\text{pl}}\xspace}}
\ex. The farmer near the clerks knows$_{\Vsg}$ many people. \label{ex:agr-pp-sg-pl-match}

\ex. *The farmer near the clerks know$_{\Vpl}$ many people. \label{ex:agr-pp-sg-pl-mismatch}

\ex. The farmers near the clerk know$_{\Vpl}$ many people. \label{ex:agr-pp-pl-sg-match}

\ex. *The farmers near the clerk knows$_{\Vsg}$ many people. \label{ex:agr-pp-pl-sg-mismatch}

\noindent one in which the distractor is in a subject-extracted
relative clause postmodifier of the subject:

\setcounter{ExNo}{0}

\ex. The farmer that embarrassed the clerks knows$_{\Vsg}$ many people. \label{ex:agr-src-sg-pl-match}

\ex. *The farmer that embarrassed the clerks know$_{\Vpl}$ many people. \label{ex:agr-src-sg-pl-mismatch}

\ex. The farmers that embarrassed the clerk know$_{\Vpl}$ many people. \label{ex:agr-src-pl-sg-match}

\ex. *The farmers that embarrassed the clerk knows$_{\Vsg}$ many people. \label{ex:agr-src-pl-sg-mismatch}

\noindent and  one in which the distractor is in an object-extracted
relative clause postmodifier of the subject:

\setcounter{ExNo}{0}

\ex. The farmer that the clerks embarrassed knows$_{\Vsg}$ many people. \label{ex:agr-orc-sg-pl-match}

\ex. *The farmer that the clerks embarrassed know$_{\Vpl}$ many people. \label{ex:agr-orc-sg-pl-mismatch}

\ex. The farmers that the clerk embarrassed know$_{\Vpl}$ many people. \label{ex:agr-orc-pl-sg-match}

\ex. *The farmers that the clerk embarrassed knows$_{\Vsg}$ many people. \label{ex:agr-orc-pl-sg-mismatch}

\paragraph{Criterion} Following \textcite{Linzen:et-al:2016} and
\textcite{Marvin:Linzen:2018}, we require successful
discrimination of the preferred upcoming verbform of the given lemma
(rather than, for example, successful discrimination of the better
context given a particular verbform).  For success we require that a
model successfully predicts the preferred verbform for \emph{both} the
singular- and plural-subject versions of an item:
\begin{align*}
  \Pcond{ex:agr-orc-sg-pl-match}(\Vsg) >
  \Pcond{ex:agr-orc-sg-pl-mismatch}(\Vpl) \wedge
  \Pcond{ex:agr-orc-pl-sg-match}(\Vpl) > \Pcond{ex:agr-orc-pl-sg-mismatch}(\Vsg) 
\end{align*}

Chance performance is thus 25\%, though a context-insensitive baseline
that places different probabilities on {\Vsg} and {\Vpl} would score 50\%.

\paragraph{References} \textcite{bock-miller:1991,Linzen:et-al:2016,Marvin:Linzen:2018}

\subsection{Reflexive pronoun licensing}
\label{sec:refl-pron-licens}

\setcounter{ExNo}{0}
\newcommand{\Rsg}{\ensuremath{{\text{R}_{\text{sg}}}}\xspace}
\newcommand{\Rsgf}{\ensuremath{{\text{R}_{\text{sg.fem}}}}\xspace}
\newcommand{\Rsgm}{\ensuremath{{\text{R}_{\text{sg.masc}}}}\xspace}
\newcommand{\Rpl}{\ensuremath{{\text{R}_{\text{pl}}}}\xspace}
\newcommand{\herself}{\ensuremath{\text{herself}_\Rsgf}\xspace}
\newcommand{\himself}{\ensuremath{\text{himself}_\Rsgm}\xspace}
\newcommand{\themselves}{\ensuremath{\text{themselves}_\Rpl}\xspace}

The noun phrase with which a reflexive pronoun (\emph{herself},
\emph{himself}, \emph{themselves}) corefers must command it in a
sense similar to that relevant for negative-polarity items
(Section~\ref{sec:negat-polar-licens}).  In the below example, the
reflexive pronoun ending the sentence can only corefer to the subject
of the sentence, \emph{author}, with which it must agree in number: a
singular subject requires a singular reflexive \Rsg, and a plural
subject requires a plural reflexive  \Rpl.

\setcounter{ExNo}{0}

\ex. The author next to the senators hurt \herself. \label{ex:reflexives-pp-fem-sg-match}

\ex. *The authors next to the senator hurt \herself. \label{ex:reflexives-pp-fem-sg-mismatch}

\ex. The authors next to the senator hurt \themselves. \label{ex:reflexives-pp-fem-pl-match}

\ex. *The authors next to the senator hurt
\themselves. \label{ex:reflexives-pp-fem-pl-mismatch}

We generated a pair of test suites---one in which the singular
reflexive is \emph{herself}, and another where the singular reflexive
is \emph{himself}, on the template of the above example, where the
distractor NP is in a prepositional-phrase postmodifier of the subject
NP.  We also generated a similar pair of test suites where the
distractor NP is inside a subject-extracted relative clause modifying
the subject:

\setcounter{ExNo}{0}

\ex. The author that  liked the senators hurt \herself. \label{ex:reflexives-src-fem-sg-match}

\ex. *The authors that  liked the senator hurt \herself. \label{ex:reflexives-src-fem-sg-mismatch}

\ex. The authors that  liked the senator hurt \themselves. \label{ex:reflexives-src-fem-pl-match}

\ex. *The authors that  liked the senator hurt
\themselves. \label{ex:reflexives-src-fem-pl-mismatch}

\noindent and a pair of test suites where the distractor NP is inside
an object-extracted relative clause modifying the subject:

\setcounter{ExNo}{0}

\ex. The author that  the senators liked hurt \herself. \label{ex:reflexives-orc-fem-sg-match}

\ex. *The authors that  the senator liked hurt \herself. \label{ex:reflexives-orc-fem-sg-mismatch}

\ex. The authors that  the senator liked hurt \themselves. \label{ex:reflexives-orc-fem-pl-match}

\ex. *The authors that   the senator liked hurt
\themselves. \label{ex:reflexives-orc-fem-pl-mismatch}

\paragraph{Criterion} For each item in each test suite, we require that
for both the singular and the plural versions of the reflexive pronoun
the model assign higher conditional probability in the correct
licensing context than in the incorrect licensing context:
\begin{align*}
  \Pcond{ex:reflexives-pp-fem-sg-match}(\Rsg) >
  \Pcond{ex:reflexives-pp-fem-sg-mismatch}(\Rsg) \wedge   \Pcond{ex:reflexives-pp-fem-pl-match}(\Rpl) >
  \Pcond{ex:reflexives-pp-fem-pl-mismatch}(\Rpl)
\end{align*}

\noindent
Chance is 25\%.
% Note that an alternative criterion would be to
% require that for both of the \emph{contexts} the model place higher
% probability 

\paragraph{References} \textcite{reinhart:1981definite,Marvin:Linzen:2018}

\subsection{Subordination}
\label{sec:subordination}

\newcommand{\nomainclause}{\ensuremath{\text{\textsc{End}}}\xspace}
\newcommand{\mainclause}{\ensuremath{\text{\textsc{MC}}}\xspace}

\setcounter{ExNo}{0}

Beginning a sentence with \emph{As}, \emph{When}, \emph{Before},
\emph{After}, or \emph{Because}, implies that an immediately following
clause is not the main clause of the sentence, as would have otherwise
been the case, but instead is a \key{subordinate clause} that must be
followed by the main clause.  Ending the sentence without a main
clause, as in \ref{ex:subordination-sub-nomc}, is problematic.
Conversely, following an initial clause with a second clause \mainclause (without
linking it to the initial clause with \emph{and}, \emph{but},
\emph{despite}, or a similar coordinator or subordinator), as in
\ref{ex:subordination-nosub-mc} below, is unexpected and odd.

\ex. The minister praised the building$\overbrace{\text{.\strut}}^{\nomainclause}$ \label{ex:subordination-nosub-nomc}

\ex. *After the minister praised the building$\overbrace{\text{.\strut}}^{\nomainclause}$ \label{ex:subordination-sub-nomc}

\ex. ??The minister praised the building$\overbrace{\text{, it started to rain.\strut}}^{\mainclause}$ \label{ex:subordination-nosub-mc}

\ex. After the minster praised the building$\overbrace{\text{, it started to
rain.\strut}}^{\mainclause}$ \label{ex:subordination-sub-mc}

In addition to the base test suite exemplified by the item above, we
include three versions with longer  and more complex initial
clauses, which may make the test suite more difficult.  In the first
of these versions, we postmodify both the subject and object of the
initial clauses with prepositional phrases:

\bigskip

\begin{minipage}[h]{0.95\linewidth}
\begin{center}
 the minister praised the building\\
  $\downarrow$\\
  the minister in the dark
suit and white tie praised the new building on the town's main square  
\end{center}
  \end{minipage}

  \bigskip
  
\noindent  In the second of these versions, the postmodifiers are
  subject-extracted relative clauses:
  
  \bigskip
  
\begin{minipage}[h]{0.95\linewidth}
  \begin{center}
the minister praised the building\\
$\downarrow$\\
the minister who wore a
black suit praised the new building that was built by the square
\end{center}
\end{minipage}

\bigskip

\noindent In the third of these versions, the postmodifiers are object-extracted
relative clauses:

\bigskip

\begin{minipage}[h]{0.95\linewidth}
    \begin{center}
      the minister praised the building\\
      $\downarrow$\\
      the minister who the mayor had invited praised the new building that the businessman had built downtown
\end{center}
  \end{minipage}

  \bigskip

% \begin{center}
%   \setlength\tabcolsep{2pt}
% \hspace{-0.25cm}  
% \begin{tabular}{p{7cm}}
%  the minister praised the building\\
%   $\downarrow$\\
%   the minister in the dark
% suit and white tie praised the new building on the town 's main square
%  \end{tabular}
% \end{center}

% In the second of these versions, the postmodifiers are
% subject-extracted relative clauses:

% \begin{center}
%   \setlength\tabcolsep{2pt}
% \hspace{-0.25cm}  
% \begin{tabular}{rp{7cm}}
% &the minister praised the building\\
%   $\rightarrow$&the minister who wore a
% black suit praised the new building that was built by the square
% \end{tabular}
% \end{center}

% In the third of these versions, the postmodifiers are object-extracted
% relative clauses:

% \begin{center}
%     \setlength\tabcolsep{2pt}
% \hspace{-0.25cm}  
% \begin{tabular}{rp{7cm}}
% &the minister praised the building\\
%   $\rightarrow$&the minister who the mayor had invited praised the new building that the businessman had built downtown
% \end{tabular}
% \end{center}

\paragraph{Criterion} Introducing a subordinator at the beginning of
the sentence should make an ending without a second clause less
probable, and should make a second clause more probable:
\begin{align*}
  \Pcond{ex:subordination-nosub-nomc}(\nomainclause) >
  \Pcond{ex:subordination-sub-nomc}(\nomainclause) \wedge  \Pcond{ex:subordination-sub-mc}(\mainclause) <
  \Pcond{ex:subordination-nosub-mc}(\mainclause)
\end{align*}

\paragraph{References} \textcite{Futrell:et-al:2018}

%%% Local Variables:
%%% mode: latex
%%% TeX-master: t
%%% End: